\documentclass[letterpaper]{article} % DO NOT CHANGE THIS
\usepackage{paper_appendix}  % DO NOT CHANGE THIS
\usepackage[hyphens]{url}  % DO NOT CHANGE THIS
\usepackage{graphicx} % DO NOT CHANGE THIS
\usepackage{natbib}  % DO NOT CHANGE THIS AND DO NOT ADD ANY OPTIONS TO IT
\usepackage{caption} % DO NOT CHANGE THIS AND DO NOT ADD ANY OPTIONS TO IT
\usepackage{algorithm}
\usepackage{algorithmic}
\usepackage{amsmath}
\usepackage{amssymb}
\usepackage{booktabs}
\usepackage{multirow}
\usepackage{makecell}
\usepackage{tabularx}
\usepackage{subcaption}
\usepackage[table,xcdraw]{xcolor}
\usepackage{enumitem}
\usepackage{ragged2e}
\usepackage{times}
\usepackage{latexsym}
\usepackage{graphicx}
\usepackage[T1]{fontenc}
\usepackage{appendix}

\definecolor{headergray}{gray}{0.90}
\definecolor{bestgray}{gray}{0.94}
\definecolor{lightblue}{HTML}{E6F0FA}
\definecolor{highlightblue}{HTML}{E6F0FA}
\newcolumntype{Y}{>{\centering\arraybackslash}X}
\usepackage{newfloat}
\usepackage{listings}
\DeclareCaptionStyle{ruled}{labelfont=normalfont,labelsep=colon,strut=off} % DO NOT CHANGE THIS
\floatstyle{ruled}
\newfloat{listing}{tb}{lst}{}
\floatname{listing}{Listing}

\usepackage{booktabs}

\title{UniFed-VLM: Federated Instruction Tuning for Vision-Language Models with Multiple Heterogeneity}
\author{
    Pengyu Wang\textsuperscript{\rm 1,2}\equalcontrib,
    Baochen Xiong\textsuperscript{\rm 1,2,3}\equalcontrib,
    Xiaoshan Yang\textsuperscript{\rm 1,2,3}\corresponding,
    Yifan Xu\textsuperscript{\rm 5},
    Zhang Qimeng\textsuperscript{\rm 4},\\
    Haifeng Chen\textsuperscript{\rm 4},
    Changsheng Xu\textsuperscript{\rm 1,2,3}
}

\affiliations{
    \textsuperscript{\rm 1}State Key Laboratory of Multimodal Artificial Intelligence Systems, 
    Institute of Automation, \\Chinese Academy of Sciences, Beijing, China\\
    \textsuperscript{\rm 2}School of Artificial Intelligence, University of Chinese Academy of Sciences, 
    Beijing, China\\
    \textsuperscript{\rm 3}Pengcheng Laboratory, Shenzhen, China\\
    \textsuperscript{\rm 4}Data Science and Artificial Intelligence Research Institute, \\China United Network Communications Group Co., Ltd., Beijing, China\\
    \textsuperscript{\rm 5}King Abdullah University of Science and Technology, Thuwal, Saudi Arabia\\
     \{wangpengyu2026, xiongbaochen2022\}@ia.ac.cn,
    \{xiaoshan.yang, csxu\}@nlpr.ia.ac.cn,
     yifan.xu@kaust.edu.sa,
    \{zhangqm50, chenhaifeng\}@chinaunicom.cn
}

\begin{document}

\maketitle

\begin{abstract}
Vision–Language Models (VLMs) have demonstrated strong performance in multimodal understanding and generation. 
However, fine-tuning of VLMs typically relies on centralized data, which raises privacy concerns in certain domains (e.g. healthcare). 
Federated Learning (FL) provides a natural solution by enabling model training without sharing raw data. However, applying FL to VLM instruction tuning is highly challenging. VLMs have substantial parameter scales, and in real-world scenarios, clients exhibit significant heterogeneity in tasks, modalities, and model architectures. Existing methods mainly focus on simplified settings and are unable to handle such multi-dimensional heterogeneous scenarios. In this work, we study federated instruction tuning under joint heterogeneity in tasks, modalities, and model architectures. We propose UniFed-VLM, a unified federated instruction tuning framework for VLMs that addresses multiple types of heterogeneity. It consists of two key components: 1) Federated Compensated Subspace Aggregation (FedCSA), which performs subspace-aligned aggregation of parameter-efficient adapters with dynamic weighting and compensation to mitigate heterogeneity-induced conflicts; 2) Two-stage Collaborative Distillation (TCoD), which enables effective knowledge transfer across heterogeneous models via a Mutual Distillation Adapter (MDA) and a mixture-of-experts-based distillation strategy. We conduct experiments on multiple benchmark datasets, and the results show that UniFed-VLM achieves stronger average performance across diverse tasks compared with existing FL methods. The source code is available at: https://github.com/wangpengyu2004/UniFed-VLM.
\end{abstract}


\begin{figure}[!th]
\centering
\centerline{\includegraphics[width=8.50cm, height=6cm]{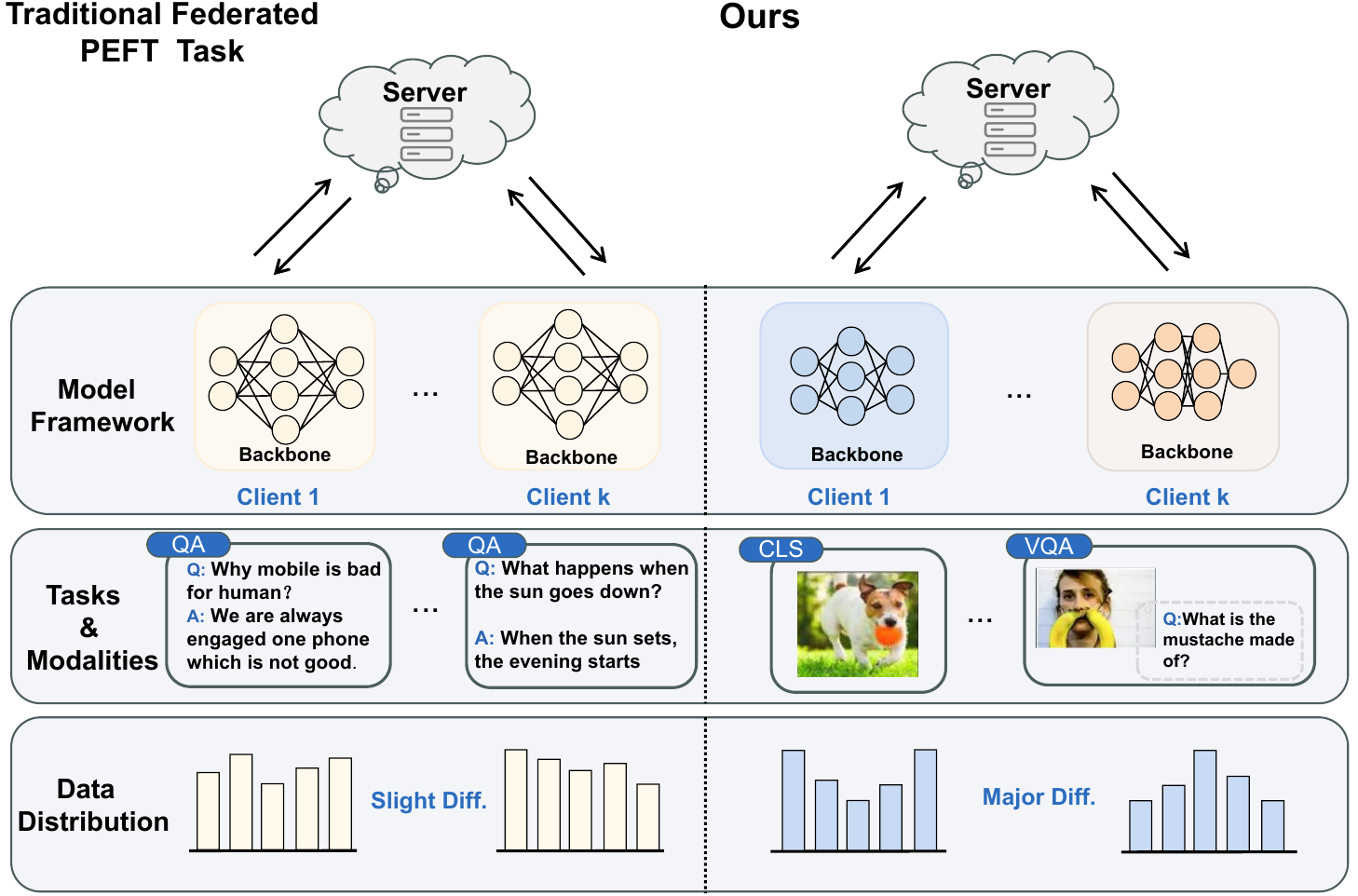}}
\caption{An overview of the federated setting with intertwined heterogeneities across models, tasks, and modalities.}
\label{fig:1}
\end{figure}

\section{Introduction}
Vision–Language Models (VLMs) \cite{xu2021vlm,achiam2023gpt}, which integrate vision and language,  have demonstrated strong cross-modal understanding and generation capabilities across various multimodal tasks, such as visual question answering (VQA) \cite{antol2015vqa}.
To adapt VLMs to downstream tasks, task-specific data are typically required for fine-tuning. 
Instruction tuning \cite{liu2023visual} and parameter-efficient tuning (PEFT) techniques \cite{ding2023parameter}, such as LoRA \cite{hu2022lora}, have been widely adopted for this purpose. However, most existing fine-tuning approaches rely on centralized training, where distributed data are collected on a central server.
In privacy-sensitive domains such as healthcare, strict regulations make centralized data aggregation impractical. 
To address this limitation, recent studies have combined federated learning \cite{yang2019federated} with instruction tuning and PEFT \cite{zhang2024towards}. Some methods build upon classical FL algorithms, such as FedAvg \cite{li2019convergence} and FedAdam \cite{reddi2020adaptive}, which aggregate client-uploaded model parameters on the server to obtain a global model without accessing local data, enabling collaborative model improvement across multiple parties while preserving data privacy.

The core challenge of federated instruction tuning or PEFT lies in client heterogeneity, which inherently hinders effective collaborative learning. Existing works have achieved promising results under various heterogeneous settings \cite{ye2023heterogeneous,yang2025survey}. For instance, HETLORA \cite{cho2024heterogeneous} considers resource and data heterogeneity, FlexLoRA \cite{bai2024federated} addresses task and resource heterogeneity, and H$^2$Tune \cite{guo2025h2tune} addresses architecture heterogeneity.
Although these approaches have yielded promising results, they typically focus on isolated or limited forms of heterogeneity.
In real-world applications, clients often exhibit simultaneous heterogeneity in modalities, tasks, and model architectures, which cannot be adequately handled by methods designed for individual types of heterogeneity.
Therefore, studying federated tuning under multi-dimensional heterogeneity remains an important yet underexplored problem.

In this paper, we consider a challenging scenario, as shown in Figure \ref{fig:1}, which simultaneously involves task, modality, and model heterogeneity. Existing methods typically address only one or two types of heterogeneity, while our setting considers all three, introducing more challenges.
This setting inherently suffers from intertwined heterogeneities spanning tasks (e.g., VQA, visual grounding, classification), modalities (e.g., text, image), and architectures (e.g., varying dimensions, layer depths, structural backbones). This joint diversity exacerbates cross-client statistical and structural discrepancies, severely obstructing knowledge transfer. Consequently, local parameter updates exhibit acute gradient and weight conflicts, rendering conventional federated aggregation strategies ineffective.

To address above challenges, we propose \textbf{UniFed-VLM}, a unified federated instruction tuning framework for VLMs that effectively handles such complex heterogeneous scenarios.
Within our framework, for parameter aggregation among clients with the same model architecture, we design \textbf{Federated Compensated Subspace Aggregation (FedCSA)}.
Specifically, SVD is used to achieve subspace alignment, and dynamic weights are computed for weighted aggregation.
In addition, an optimization problem is introduced to obtain complementary parameters, reducing knowledge loss during aggregation.
This method alleviates conflicts caused by task and modality heterogeneity, enabling the global parameters to capture task-agnostic shared knowledge and thus maintain strong performance across multiple tasks.
Building on this, we further propose the \textbf{Two-stage Collaborative Distillation (TCoD)}, where each client is equipped with a \textbf{Mutual Distillation Adapter (MDA)} for mutual knowledge distillation (MKD) \cite{gou2021knowledge}.
TCoD performs bidirectional knowledge transfer through two stages: first distilling knowledge from the local VLM to the MDA, and then transferring the acquired knowledge back from the MDA to the VLM with cross-model interaction.
This enables effective distillation with the local VLM, allowing it to acquire knowledge from heterogeneous models. The proposed framework achieves promising results across multiple datasets.

Our contributions can be summarized as follows:
\begin{itemize}
\item To the best of our knowledge, we are the first to explore federated instruction tuning under simultaneous task, modality, and model heterogeneity.
\item We propose UniFed-VLM, a unified framework engineered to address multi-dimensional heterogeneities in federated instruction tuning. UniFed-VLM incorporates Federated Compensated Subspace Aggregation to reconcile task and modality discrepancies, alongside Two-stage Collaborative Distillation to facilitate seamless knowledge transfer across heterogeneous architectures.
\item Extensive experiments on two public datasets demonstrate the effectiveness of UniFed-VLM.
\end{itemize}

\section{Related Work}

\textbf{Vision-Language Models}.
VLMs integrate multimodal information, such as vision and text, to achieve strong multimodal capabilities across downstream tasks. VLMs can be broadly categorized from two perspectives. Based on modality fusion, they are divided into dual-stream architectures \cite{lu2023fedclip,li2020oscar,lu2019vilbert}, which separately encode visual and textual inputs before cross-modal fusion, such as CLIP \cite{radford2021learning}, and single-stream architectures \cite{achiam2023gpt,kim2021vilt,su2019vl,li2020unicoder}, which jointly process multimodal inputs using a unified encoder. Based on visual encoding, VLMs can also be classified as encoder-based models, which employ a visual encoder followed by an input projector to align visual and language features before integrating with LLMs, such as LLaVA \cite{liu2023visual} and Flamingo \cite{alayrac2022flamingo}, or encoder-free models, which rely on a unified Transformer backbone without a separate visual encoder, such as Show-O \cite{xie2024show}. In this work, we explicitly consider these architectural differences.

\begin{figure*}[htb]
  \centering
  \includegraphics[width=16.3cm, height=8.85cm]{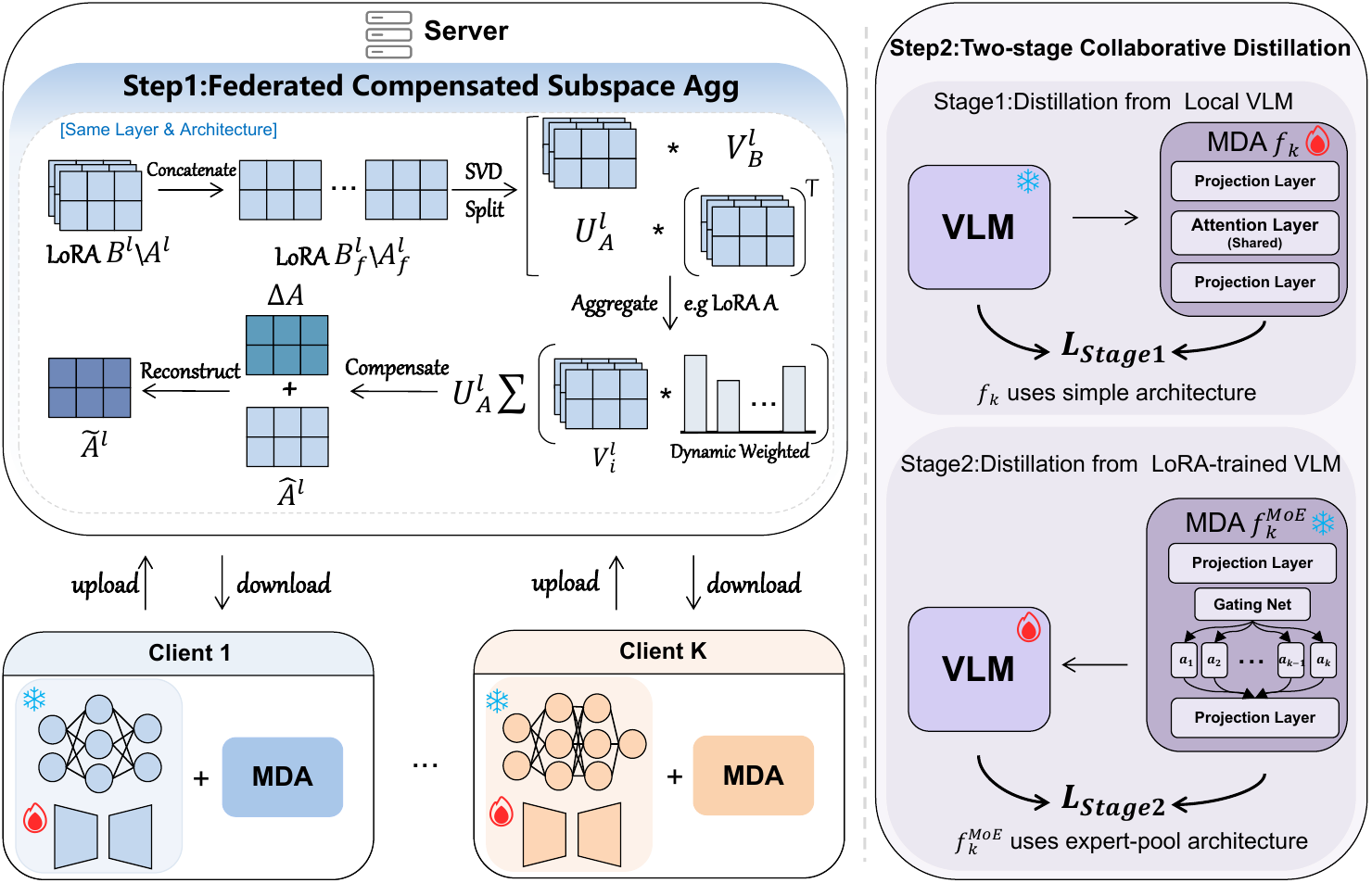} 
  \caption{\textbf{Overview of UniFed-VLM.} Each client consists of a local VLM and an MDA. 
The training involves two steps. In Step 1, homogeneous LoRA parameters from the same model group are aggregated via FedCSA with SVD-based subspace alignment, dynamic weighting, and compensation optimization. 
In Step 2, the aggregated parameters are broadcast to clients, where the local VLM and MDA perform two-stage knowledge distillation with different MDA architectures across stages.}
  \label{fig:2}
\end{figure*}

\textbf{Federated Learning}.
The core idea of federated learning (FL) is to enable decentralized collaborative training while preserving data privacy and local autonomy. Classical FL algorithms, such as FedAvg and its variants FedAdam and FedProx \cite{reddi2020adaptive}, address non-IID data distributions through iterative parameter aggregation. Recently, FL has been extended to foundation model fine-tuning, where parameter-efficient fine-tuning (PEFT) is adopted to reduce communication and computational costs. FedDAT \cite{chen2024feddat} introduces a Dual-Adapter Teacher with bidirectional knowledge distillation to alleviate task conflicts and data heterogeneity. FLORA \cite{wang2024flora} assigns different LoRA ranks to clients with varying resources for heterogeneous low-rank adaptation. 
FlexLoRA \cite{bai2024federated} leverages SVD to dynamically allocate heterogeneous LoRA modules based on client resources and task characteristics, mitigating task and resource heterogeneity.

Model heterogeneity has also been explored in FL. FedConv \cite{shen2024fedconv} compresses global parameters into heterogeneous models via convolutional layers and decompresses them for aggregation, but is mainly applicable to small-scale models. FedMD \cite{li2019fedmd} bridges heterogeneous models through knowledge distillation with a server-maintained public dataset, introducing additional computational and storage costs. FedProto \cite{tan2022fedproto} and FedTGP \cite{zhang2024fedtgp} reduce this burden by enabling clients to upload local prototypes while the server maintains global prototypes for local guidance. For foundation model heterogeneity, pFedLoRA \cite{yi2023pfedlora} introduces a model-heterogeneous personalized federated learning framework based on LoRA tuning, where homogeneous adapters are used as knowledge carriers to enable knowledge exchange among heterogeneous local models. H$^2$Tune \cite{guo2025h2tune} constructs a shared global matrix with a unified rank for heterogeneous LoRA modules, enabling knowledge aggregation across different LoRA structures. HeteroTune \cite{jia2024heterotune} introduces Dense Mixture of Adapters with Cross-Model Gradient Alignment (CMGA) to address model heterogeneity, but only considers feature-dimensional differences.
Although these methods address specific heterogeneity scenarios, they do not explicitly handle the coupled challenges of simultaneous task, modality, and model heterogeneity.

\section{Methodology}
\subsection{Problem Statement}
In our work, we consider a federated learning setting with multi-dimensional heterogeneity across models, tasks, and modalities. Specifically, we consider $K$ clients, each with a private dataset $D_k$, a VLM $M_k$, and a task $T_k$. For any two clients $i,j\in{1,\ldots,K}$, task heterogeneity may exist ($T_i\neq T_j$), which often induces modality heterogeneity ($m_i\neq m_j$); for example, classification and VQA involve different input-output modalities. Data sizes are also imbalanced across tasks and clients ($n_i\neq n_j$). Meanwhile, client models can differ in the number of layers ($L_i\neq L_j$), feature dimensions ($d_i\neq d_j$), architectures ($F_i\neq F_j$), and LoRA ranks ($r_i\neq r_j$). Our goal is to leverage parameter-efficient fine-tuning (PEFT) in federated learning to learn a set of heterogeneous global models $f_{\theta}$, each achieving strong performance on its corresponding client data.

\subsection{Framework Overview}
We first present an overview of \textbf{UniFed-VLM}. As illustrated in Figure~\ref{fig:2}, the framework consists of two core modules: \textbf{Federated Compensated Subspace Aggregation (FedCSA)} and \textbf{Two-stage Collaborative Distillation (TCoD)}. In Step 1, FedCSA aggregates clients sharing the same model architecture. For each layer, LoRA parameters $A_k^l\in\mathbb{R}^{r\times d_{\text{in}}}$ and $B_k^l\in\mathbb{R}^{d_{\text{out}}\times r}$ are concatenated across clients into $A_f^l\in\mathbb{R}^{r\times Nd_{\text{in}}}$ and $B_f^l\in\mathbb{R}^{Nd_{\text{out}}\times r}$, where $N$ denotes the number of clients with the same model architecture. SVD is then applied to the concatenated matrices. Based on parameter changes between consecutive rounds, dynamic weights $W_k^l$ are computed to aggregate task feature vectors $V_k^l\in\mathbb{R}^{r\times d_{\text{in}}}$ and $U_k^l\in\mathbb{R}^{d_{\text{out}}\times r}$ obtained by partitioning the singular vectors. An additional optimization step learns complementary parameters $\Delta A^l\in\mathbb{R}^{r\times d_{\text{in}}}$ and $\Delta B^l\in\mathbb{R}^{d_{\text{out}}\times r}$ to mitigate knowledge loss during aggregation. In Step 2, TCoD equips each client with an adapter for mutual knowledge distillation, with two variants corresponding to different training stages. The adapter comprises globally shared attention layers $L_{\text{att}}^k$ and client-specific projection layers $L_{\text{in}}^k$ and $L_{\text{out}}^k$, enabling bidirectional knowledge distillation through two-stage training.

\subsection{Federated Compensated Subspace Aggregation}
To mitigate the conflicts arising from task and modality heterogeneity during parameter aggregation among clients with the same model architecture, we focus on aggregating LoRA parameters under the PEFT setting, where backbone parameters of VLMs are frozen and only LoRA parameters are trained. For the $l$-th layer and the $k$-th client, the LoRA parameters are denoted as $A_k^l$ and $B_k^l$.Let $i = \{1, \dots, N\}$ denote the set of clients sharing the same model architecture. We concatenate the LoRA parameters $A_i^l$ and $B_i^l$ across clients along the non-rank dimensions. Specifically,
\begin{equation}
A_f^l=[A_1^l,\ldots,A_N^l]\in\mathbb{R}^{r\times Nd_{\text{in}}},\quad
B_f^l=
\begin{bmatrix}
B_1^l\\
\vdots\\
B_N^l
\end{bmatrix}
\in\mathbb{R}^{Nd_{\text{out}}\times r}.
\end{equation}
We then apply randomized SVD to the concatenated matrices:
\begin{equation}
A_f^l = U_A^l \Sigma_A^l (V_A^l)^\top, \qquad B_f^l = U_B^l \Sigma_B^l (V_B^l)^\top.
\end{equation}

Based on the concatenation structure, we partition the right singular matrix $V_A^l$ and the left singular matrix $U_B^l$ to obtain task-specific feature vectors:
\begin{equation}
V_A^l=[(V_1^l)^\top,\ldots,(V_N^l)^\top]^\top,\quad
V_i^l\in\mathbb{R}^{r\times d_{\text{in}}}.
\end{equation}
\begin{equation}
U_B^l = [U_1^l, U_2^l, \dots, U_N^l], \quad U_i^l \in \mathbb{R}^{d_{\text{out}} \times r}.
\end{equation}

Since the task feature vectors (e.g., $V_1^l$ and $V_2^l$) share a common orthogonal basis, they can be aligned within a unified representation space where task-specific variations are encoded along consistent directions~\cite{stoica2025knots}. Such alignment reduces interference during aggregation and alleviates the conflicts introduced by task and modality heterogeneity.

To handle the varying quality of client updates, we introduce a dynamic weighting mechanism based on update variability. At round $t$:
\begin{equation}
\Delta A_i^{l,(t)} = A_i^{l,(t)} - A_i^{l,(t-1)},   \Delta B_i^{l,(t)} = B_i^{l,(t)} - B_i^{l,(t-1)}.
\end{equation}
The update magnitudes are defined as:
\begin{equation}
s_{A,i}^{l,(t)} = \|\Delta A_i^{l,(t)}\|_F, \qquad s_{B,i}^{l,(t)} = \|\Delta B_i^{l,(t)}\|_F.
\end{equation}
We normalize these values and apply momentum updates($\beta$ is a hyperparameter, typically set to 0.99),such as:
\begin{equation}
v_{A,i}^{l,(t)} = \beta v_{A,i}^{l,(t-1)} + (1-\beta) \frac{s_{A,i}^{l,(t)}}{\max_j s_{A,j}^{l,(t)} + \epsilon}.
\end{equation}
The final aggregation weights are defined as:
\begin{equation}
w_{A,i}^l = \frac{1/(v_{A,i}^{l,(t)} + \epsilon)}{\sum_j 1/(v_{A,j}^{l,(t)} + \epsilon)}.
\end{equation}
Clients with smaller update variations receive higher weights, since stable updates are more likely to preserve shared knowledge, whereas large fluctuations often indicate task-specific deviations under heterogeneous objectives. The aggregated subspace representations are:
\begin{equation}
\widetilde{V}^l = \sum_i w_{A,i}^l V_i^l, \qquad \widetilde{U}^l = \sum_i w_{B,i}^l U_i^l.
\end{equation}
We then reconstruct the aggregated parameters:
\begin{equation}
\hat{A}^l = U_A^l \Sigma_A^l (\widetilde{V}^l)^\top, \qquad \hat{B}^l = \widetilde{U}^l \Sigma_B^l V_B^l.
\end{equation}

This can be viewed as weighted aggregation in a geometrically aligned shared subspace, which preserves consistent semantic structures better than direct parameter averaging.

We aggregate $A_i^l$ and $B_i^l$ separately. Compared with aggregating the full matrix $\Delta W_i^l=B_i^lA_i^l$, this is more efficient and memory-friendly, but may introduce information loss. To compensate for this, we learn $\Delta A^l$ and $\Delta B^l$, and denote $\tilde{A}^l=\hat{A}^l+\Delta A^l$ and $\tilde{B}^l=\hat{B}^l+\Delta B^l$. The objective is
\begin{equation}
\mathcal{L}_{\text{comp}}
=
\sum_{l=1}^{L}
\sum_{i=1}^{N}
\left[
\left\langle
-B_i^l A_i^l,
B_i^l A_i^l-\tilde{B}^l\tilde{A}^l
\right\rangle
\right]^2 .
\end{equation}
This encourages the compensated global parameters to approximate the original client mappings and is optimized with Adam for a few hundred steps. The procedure acts as low-rank error recovery, mitigating the information loss introduced by SVD-based aggregation.

\subsection{Two-stage Collaborative Distillation}
After aggregating clients with the same architecture, we further enhance the global model through knowledge interaction across heterogeneous models. Unlike conventional distillation methods that require server-side shared data or class prototypes, we attach a Mutual Distillation Adapter (MDA) to each client. Only its core attention parameters are uploaded, enabling efficient knowledge transfer without parameter aggregation.

\textbf{Stage 1: Distillation from VLM to MDA}. In the first stage, we freeze the VLM and distill knowledge into a MDA $f_k$. The MDA is defined as:
\begin{equation}
f_k(x) = L_{\text{out}}^k \left( L_{\text{att}}^k \left( L_{\text{in}}^k (x) \right) \right).
\end{equation}
Here, $L_{\text{in}}^k$ and $L_{\text{out}}^k$ are client-specific projection layers. $L_{\text{in}}^k$ consists of linear layers that map the input from dimension $d_{\text{in}}$ to a shared hidden dimension $d_h$, while $L_{\text{out}}^k$ symmetrically maps the hidden representation to the output space $d_{\text{out}}$. The input/output dimensions depend on the client.
The attention module $L_{\text{att}}^k$ is shared across all clients and consists of multiple Transformer layers with a unified hidden dimension $d_h$. This stage enables the MDA $f_k$ to capture client-specific knowledge from the corresponding VLM. We first distill knowledge from the VLM into the MDA for several epochs, enabling it to acquire the knowledge of the original VLM.

\textbf{Stage 2: Distillation from MDA to VLM}. In the second stage, knowledge is distilled back from the MDA to the VLM. The adapter, denoted as $f_k^{\text{MoE}}$, differs from that in Stage 1 by introducing a Mixture-of-Experts (MoE) mechanism. Besides the original attention module $L_{\text{att}}^k$, it includes $M$ attention experts $\{E_i\}_{i=1}^{M}$, where $M$ is the number of clients with heterogeneous model architectures and each expert corresponds to the core attention module of a heterogeneous client. A gating network performs expert routing. During this stage, only the gating network in the MDA and the LoRA parameters are trainable, while all other parameters remain frozen. We further introduce the task loss $\mathcal{L}{\text{task}}^{\text{MoE}}$ and the MoE balancing loss $\mathcal{L}_{\text{Bal}}^{\text{MoE}}$~\cite{wang2024auxiliary} to jointly optimize the gating network and the VLM.

Given an input $x$, we first obtain the hidden input $h$ via processing through $L_{\text{in}}^k$. Then, the gating network computes expert weights:
\begin{equation}
g = \text{Softmax} \left( W_g \cdot \text{Pool}(h) \right).
\end{equation}
We then select the Top-$K$ experts and compute:
\begin{equation}
h_{\text{MoE}} = \sum_{i \in \text{TopK}(g)} \tilde{g}_i \cdot E_i(h),
\end{equation}
where $\tilde{g}_i$ denotes the normalized weights.

To improve training stability, we treat the original attention module of the client as an anchor expert and fuse it with the MoE output:
\begin{equation}
h_{\text{final}} = \alpha h_{\text{MoE}} + (1-\alpha) E_{\text{anchor}}(h),
\end{equation}
where $\alpha \in [0, 1]$ is a balancing coefficient. Finally, we obtain the final output by transforming $h_{\text{final}}$ through $L_{\text{out}}^k$.This design stabilizes training while preserving a reliable base representation.

Considering that many tokens in sequence modeling are uninformative, we perform distillation only on valid tokens. The distillation loss is defined as:
\begin{equation}
\mathcal{L}_{\text{KD}} = \frac{1}{|M|} \sum_{i \in M} D_{\text{KL}} \left( p_i^{(s)} \parallel p_i^{(t)} \right),
\end{equation}
where $M$ denotes the set of valid non-mask tokens (excluding tokens with label -100) in the sequence, $p_i^{(s)}$ and $p_i^{(t)}$ represent the student and teacher predictions at the $i$-th token position, respectively.

The overall objectives for the two stages are:
\begin{equation}
\begin{split}
\mathcal{L}_{\text{Stage1}} &= \mathcal{L}_{\text{task}}^l + \lambda_{\text{S1}} \mathcal{L}_{\text{KD}} \left( p_l \parallel p_V \right), \\
\mathcal{L}_{\text{Stage2}} &= \mathcal{L}_{\text{task}}^V + \lambda_{\text{S2}} \mathcal{L}_{\text{KD}} \left( p_V \parallel p_l \right) + \lambda_{\text{task}}^{\text{MoE}} \mathcal{L}_{\text{task}}^{\text{MoE}}+\lambda_{\text{MoE}} \mathcal{L}_{\text{Bal}}^{\text{MoE}}.
\end{split}
\end{equation}
Here, $\mathcal{L}_{\text{task}}^l$ and $\mathcal{L}_{\text{task}}^V$ denote the task losses for the MDA and the VLM, respectively (both implemented as autoregressive losses). The two KL terms correspond to the distillation directions in the two stages.The hyperparameters $\lambda_{\text{S1}}$,$\lambda_{\text{S2}}$,$\lambda_{\text{task}}^{\text{MoE}} $ and $\lambda_{\text{MoE}}$ control the relative weights of different loss terms in the objective function.

\begin{table*}[t]
\centering
\small
\setlength{\tabcolsep}{3pt} 
\begin{tabularx}{\textwidth}{llYYYYYYYY}
\toprule
\multirow{3}{*}{\textbf{Scenario}} &
\multirow{3}{*}{\textbf{Method}}
& \multicolumn{4}{c}{\textbf{Fed-Nature}}
& \multicolumn{4}{c}{\textbf{Fed-Crossdomain}} \\
\cmidrule(lr){3-6}\cmidrule(lr){7-10}

& & VQA & Caption & VG & CLS
& VQA & Caption & VG & CLS \\

& & Acc$\uparrow$ & CIDEr/R-L$\uparrow$ & IoU$\uparrow$ & Acc$\uparrow$
& Acc$\uparrow$ & CIDEr/R-L$\uparrow$ & IoU$\uparrow$ & Acc$\uparrow$ \\
\midrule

\multirow{8}{*}{Scenario 1}
& LOCAL     & 0.712 & 0.705/0.348 & 0.390 & 0.891 & 0.509 & 2.168/0.569 & 0.311 & 0.213 \\
& FedAdam   & 0.709 & 0.763/0.361 & 0.499 & 0.910 & 0.541 & 1.201/0.527 & 0.319 & 0.222 \\
& FedAvg    & 0.734 & 0.744/0.349 & 0.465 & 0.912 & 0.542 & 2.501/0.624 & 0.378 & 0.226 \\
& FedProx   & 0.702 & 0.753/0.357 & 0.457 & 0.906 & 0.550 & 1.748/0.579 & 0.367 & 0.234 \\
& FlexLoRA  & 0.693 & 0.729/0.357 & 0.423 & 0.893 & 0.553 & \textbf{2.994/0.651} & 0.351 & 0.227 \\
& HetLoRA   & 0.714 & 0.726/0.341 & 0.511 & 0.917 & 0.529 & 1.853/0.586 & 0.356 & 0.228 \\
& pFedLoRA  & 0.588 & 0.675/0.331 & 0.410 & 0.896 & 0.510 & 0.498/0.378 & \textbf{0.470} & 0.224 \\
\rowcolor{lightblue}& \textbf{Ours} & \textbf{0.743} & \textbf{0.784/0.358} & \textbf{0.519} & \textbf{0.918} & \textbf{0.568} & 2.786/0.634 & 0.365 & \textbf{0.237} \\
\rowcolor{bestgray}&  Central   & 0.755 & 0.863/0.361 & 0.475 & 0.913 & 0.628 & 2.936/0.647 & 0.435 & 0.321 \\

\midrule

\multirow{8}{*}{Scenario 2}
& LOCAL     & 0.677 & 0.676/0.357 & 0.335 & 0.886 & 0.511 & 2.175/0.573 & 0.220 & 0.223 \\
& FedAdam   & 0.661 & 0.410/0.290 & 0.294 & 0.878 & 0.468 & 1.587/0.537 & 0.219 & 0.228 \\
& FedAvg    & 0.716 & 0.723/0.352 & 0.341 & 0.874 & 0.684 & 1.892/0.589 & 0.194 & 0.226 \\
& FedProx   & 0.716 & 0.721/0.349 & 0.362 & 0.906 & 0.681 & 2.299/0.613 & 0.203 & 0.226 \\
& FlexLoRA  & 0.720 & 0.726/0.357 & 0.340 & 0.909 & 0.627 & 2.400/0.621 & 0.184 & 0.225 \\
& HetLoRA   & 0.739 & \textbf{0.747/0.351} & 0.347 & 0.910 & 0.639 & 2.081/0.602 & 0.208 & 0.228 \\
& pFedLoRA  & 0.700 & 0.697/0.337 & 0.299 & 0.908 & 0.327 & 0.499/0.375 & \textbf{0.319} & 0.226 \\
\rowcolor{highlightblue}& \textbf{Ours} & \textbf{0.762} & 0.713/0.346 & \textbf{0.364} & \textbf{0.917} & \textbf{0.691} & \textbf{2.678/0.640} & 0.232 & \textbf{0.239} \\
\rowcolor{bestgray}&  Central   & 0.752 & 0.863/0.356 & 0.405 & 0.913 & 0.696 & 2.936/0.647 & 0.385 & 0.321 \\
\bottomrule
\end{tabularx}
\caption{Comparison of LOCAL, Centralized, and federated baselines with UniFed-VLM on the Fed-Nature and Fed-Crossdomain datasets under different heterogeneous settings.}
\label{tab:main}
\end{table*}

\section{Experiments}
\subsection{Settings}
\textbf{Datasets and Models}. We conduct experiments on two multimodal instruction tuning datasets from FedVLMBench \cite{zheng2025fedvlmbench} : Fed-Nature and Fed-Crossdomain. Specifically, the Fed-Nature dataset is entirely derived from COCO \cite{lin2014microsoft} and includes four tasks: VQA, visual grounding, caption generation, and classification. Each task is assigned to one client, with 5,000 training samples and 1,000 test samples.The Fed-Crossdomain dataset is constructed from multiple datasets (including Fed-FGVC, Fed-ScienceCap, Fed-SLAKE, and Fed-RadGenome). It contains the same set of tasks as Fed-Nature; however, each task is assigned to three clients. The total number of samples varies across tasks, and the data distributions among clients within the same task are non-identical (non-IID). For the model design, we consider two types of models with completely different architectures and layer configurations. Specifically, we adopt:1) the architecture of LLaVA 1.5, which consists of a pre-trained CLIP visual encoder (ViT-B/32) \cite{dosovitskiy2020image} and LLaMA 3.2-3B \cite{meta2024introducing} ;2) the Show-O 1.5B model initialized with pre-trained weights.

\textbf{Baseline}. We evaluate UniFed-VLM under two heterogeneous scenarios. In Scenario 1, LLaVA is trained on classification and VQA, while Show-O is trained on caption generation and visual grounding. In Scenario 2, LLaVA is trained on classification and visual grounding, while Show-O is trained on caption generation and VQA.
We compare with LOCAL, several representative federated baselines, and a centralized upper bound. Specifically, LOCAL independently trains each client without federated operations (e.g., parameter aggregation), serving as the lower bound. The federated baselines include FedAvg, FedAdam, HetLoRA, pFedLoRA and FlexLoRA, all without cross-model techniques. We further fine-tune the model on the union of all client datasets (i.e., centralized training) and report its performance as the upper bound.

\textbf{Implementation Details}. All experiments are conducted using the PyTorch framework on four A100 40GB GPUs. We perform a total of 20 communication rounds. The Adam optimizer is used with an initial learning rate of 1e-4, which is gradually decayed to 1e-6 over rounds. For optimizing the supplementary parameters, we also use the Adam optimizer with a maximum of 200 optimization steps. The MDA are configured based on the specific dimensions and layer structures of the VLMs. Finally, we evaluate performance using task-specific metrics as well as communication overhead. Additional implementation details are provided in the supplementary material.

\begin{table}[t]
\centering
\small
\setlength{\tabcolsep}{4pt}
\begin{tabular*}{\columnwidth}{@{\extracolsep{\fill}}lcccc}
\toprule
\multirow{2}{*}{\textbf{Method}} & \textbf{VQA} & \textbf{Caption} & \textbf{VG} & \textbf{CLS} \\
\cmidrule(lr){2-5}
 & Acc$\uparrow$ & CIDEr/R-L$\uparrow$ & IoU$\uparrow$ & Acc$\uparrow$ \\
\midrule
\rowcolor{lightblue}
\textbf{UniFed-VLM} & \textbf{0.743} & \textbf{0.784/0.358} & \textbf{0.519} & \textbf{0.918} \\
w/o FedCSA (w/ FedAvg) & 0.708 & 0.746/0.361 & 0.445 & 0.907 \\
w/o FedCSA (w/ FedProx) & 0.711 & 0.781/0.365 & 0.484 & 0.901 \\
w/o TCoD & 0.707 & 0.766/0.361 & 0.458 & 0.911 \\
w/o Compensation & 0.726 & 0.774/0.357 & 0.503 & 0.915 \\
w/o MoE & 0.723 & 0.762/0.361 & 0.422 & 0.899 \\
w/o Dynamic Weighting & 0.702 & 0.763/0.360 & 0.457 & 0.905 \\
\bottomrule
\end{tabular*}
\caption{Ablation study on UniFed-VLM. ''w/'' indicates replacing FedCSA with an alternative aggregation method.}
\label{tab:ablation}
\end{table}

\subsection{Main Results}
We evaluate UniFed-VLM under two heterogeneous settings (Scenario 1 and Scenario 2). The overall results are summarized in Table~\ref{tab:main}. Compared with the baselines, UniFed-VLM achieves better average performance. In Scenario 1 on the Fed-Nature dataset, it improves performance across all tasks. For example, it outperforms FlexLoRA by 2.5\% on classification and most baselines by approximately 4\% on VQA.
This advantage is further observed under more challenging heterogeneous settings. On the Fed-Crossmain dataset, the heterogeneity is more pronounced due to its cross-domain nature and the non-IID distribution across clients. For example, in Scenario 1, although pFedLoRA achieves an IoU of 0.470 on visual grounding, its performance on the other tasks is considerably lower (e.g., 0.510 VQA accuracy and 0.498 CIDEr), indicating that it favors specific tasks under heterogeneous settings. In contrast, UniFed-VLM maintains more balanced performance across heterogeneous tasks. A similar trend is observed in Scenario 2.
In addition, UniFed-VLM approaches or even surpasses the centralized upper bound on some individual tasks. For example, in Scenario 1, it achieves 91.8\% classification accuracy on Fed-Nature, compared with 91.3\% for centralized training. Overall, these results indicate that UniFed-VLM is effective for federated instruction tuning under multi-dimensional heterogeneity.

\subsection{Ablation Study}
To evaluate the contribution of FedCSA, we replace it with the classical federated learning algorithms FedAvg and FedProx. Both replacements lead to consistent performance drops across all tasks. In particular, the VQA accuracy decreases by 3.5\% and 3.2\%, respectively, while the classification accuracy decreases by 1.1\% and 1.7\%, demonstrating that FedCSA effectively mitigates conflicts caused by task heterogeneity. Removing the compensation mechanism or dynamic weighting strategy within FedCSA also reduces performance, indicating that the compensation parameters help recover information lost during low-rank aggregation, while dynamic weighting improves aggregation under heterogeneous client updates.
We further evaluate TCoD by removing it from the framework. Without cross-model knowledge transfer, the VQA accuracy decreases by 3.6\% and the visual grounding IoU by 0.061, demonstrating the importance of TCoD for heterogeneous knowledge transfer. Replacing the MoE module with a single expert also degrades performance across all tasks, suggesting the benefit of expert routing for cross-model distillation.
Overall, the ablation results indicate that both FedCSA and TCoD, together with their key components, contribute to the performance of UniFed-VLM.

\begin{figure*}[t]
\centering
\begin{subfigure}[t]{0.245\textwidth}
    \centering
    \includegraphics[width=\linewidth]{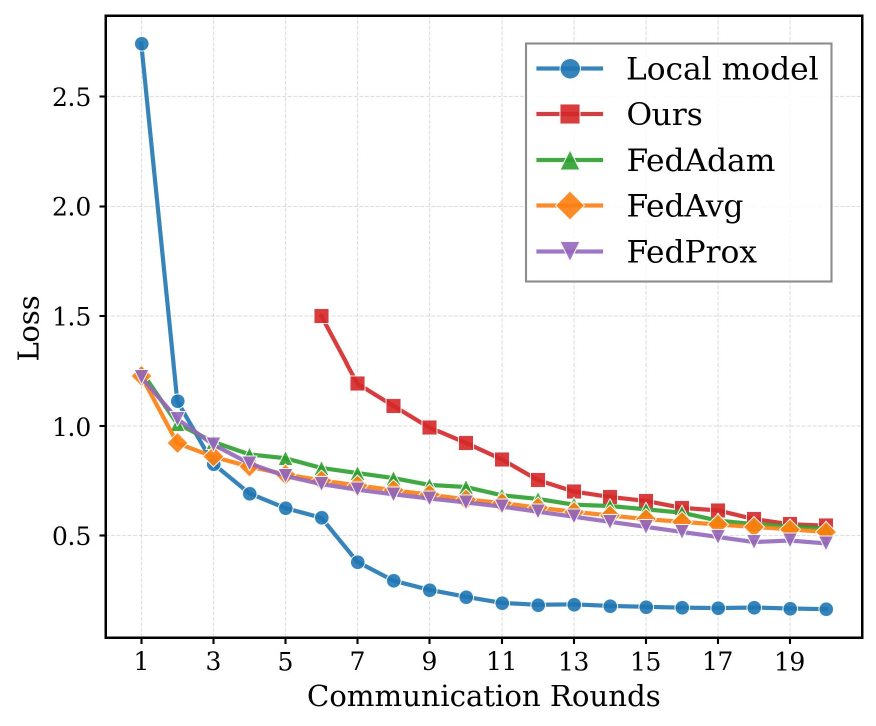}
    \caption{}
    \label{fig:loss}
\end{subfigure}\hfil
\begin{subfigure}[t]{0.245\textwidth}
    \centering
    \includegraphics[width=\linewidth]{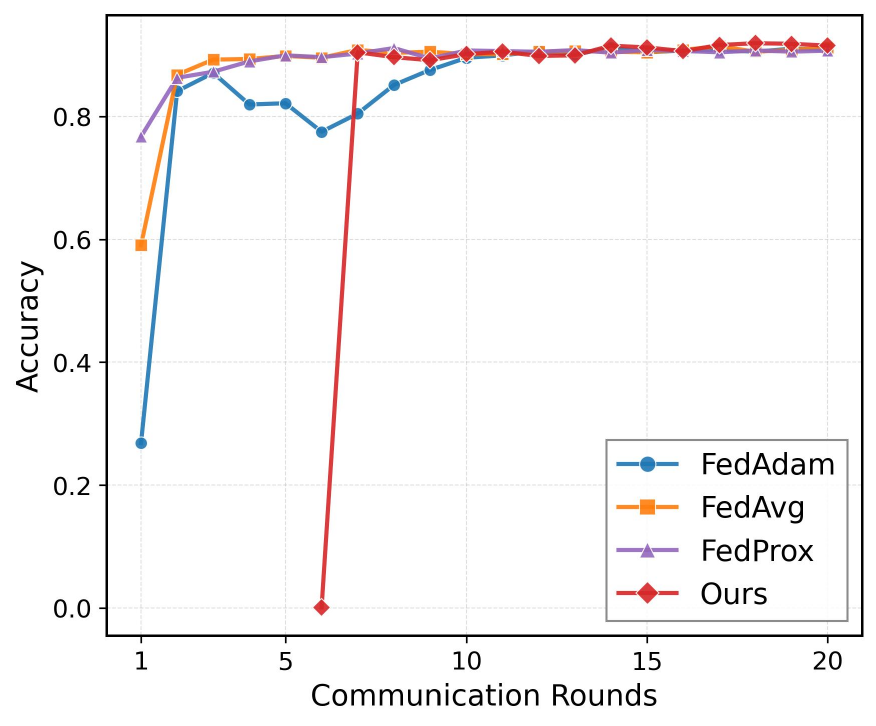}
    \caption{}
    \label{fig:cls}
\end{subfigure}\hfil
\begin{subfigure}[t]{0.245\textwidth}
    \centering
    \includegraphics[width=\linewidth]{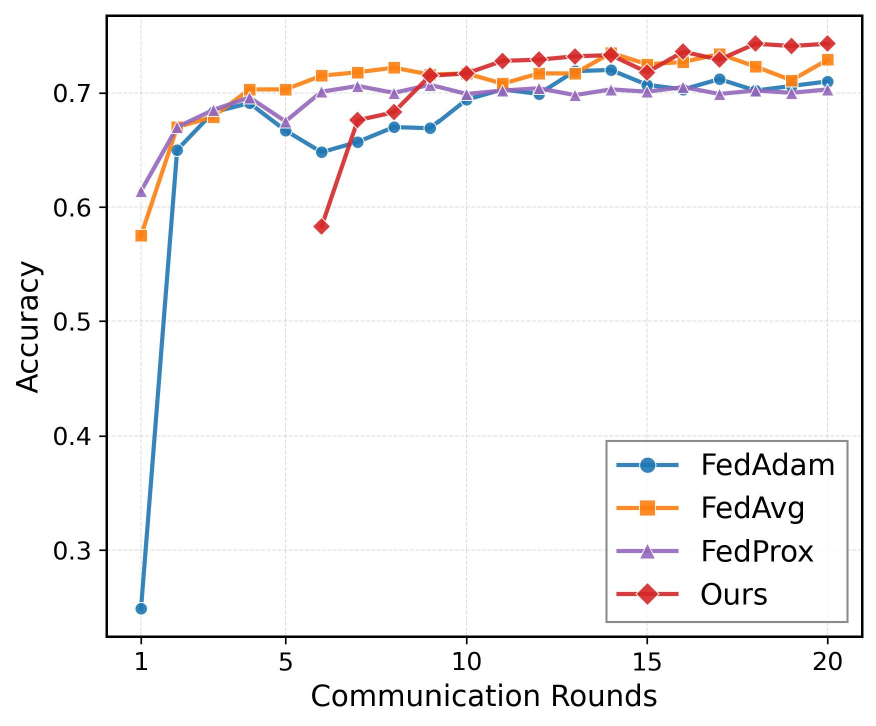}
    \caption{}
    \label{fig:vqa}
\end{subfigure}\hfil
\begin{subfigure}[t]{0.245\textwidth}
    \centering
    \includegraphics[width=\linewidth]{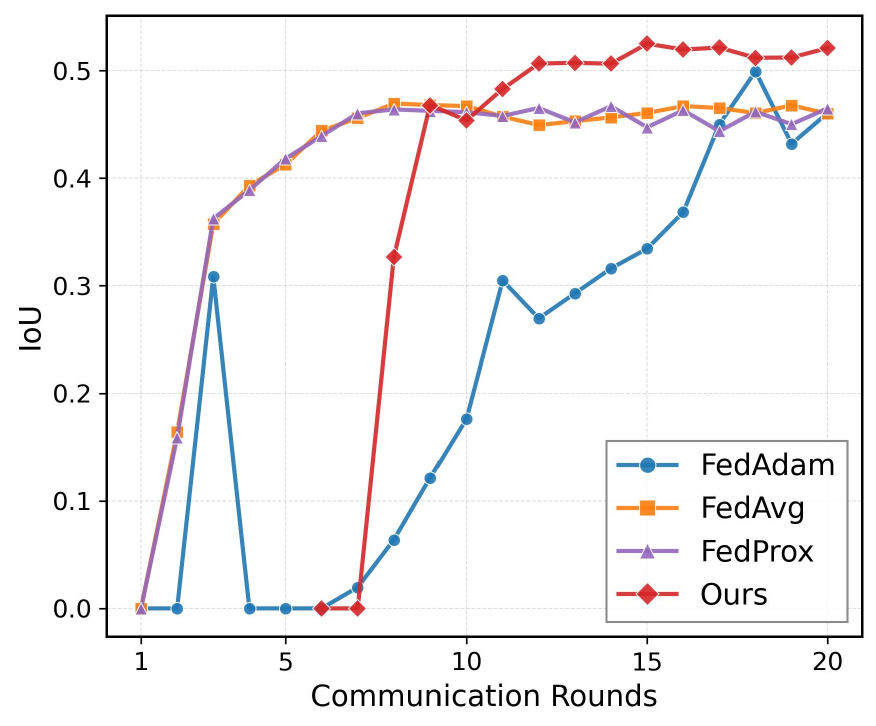}
    \caption{}
    \label{fig:vg}
\end{subfigure}

\caption{Convergence and performance comparisons on Fed-Nature under Scenario 1. (a) Average training loss across all tasks (including the MDA module). (b)–(d) Performance on classification, VQA, and visual grounding, respectively.}
\label{fig:convergence}
\end{figure*}
    
\begin{table}[t]
\centering
\small
\setlength{\tabcolsep}{4pt}
\begin{tabular*}{\columnwidth}{@{\extracolsep{\fill}}lcccc}
\toprule
\multirow{2}{*}{\textbf{LoRA Rank}} & \textbf{VQA} & \textbf{Caption} & \textbf{VG} & \textbf{CLS} \\
\cmidrule(lr){2-5}
 & Acc$\uparrow$ & CIDEr/R-L$\uparrow$ & IoU$\uparrow$ & Acc$\uparrow$ \\
\midrule
8 / 8 & \textbf{0.743} & 0.784/0.358 & 0.519 & 0.918 \\
16 / 16 & 0.732 & 0.805/0.365 & \textbf{0.533} & 0.915 \\
32 / 32 & 0.736 & \textbf{0.824/0.371} & 0.522 & \textbf{0.922} \\
\cmidrule{1-5}
16 / 8 & 0.717 & 0.792/0.367 & 0.492 & 0.915 \\
8 / 16 & 0.712 & 0.756/0.360 & 0.487 & 0.913 \\
\bottomrule
\end{tabular*}
\caption{Effect of homogeneous and heterogeneous LoRA ranks on the Fed-Nature dataset (Scenario 1).}
\label{tab:lora_rank}
\end{table}

\begin{figure}[t]
    \centering
    \begin{subfigure}{0.48\linewidth}
        \centering
        \includegraphics[width=\linewidth]{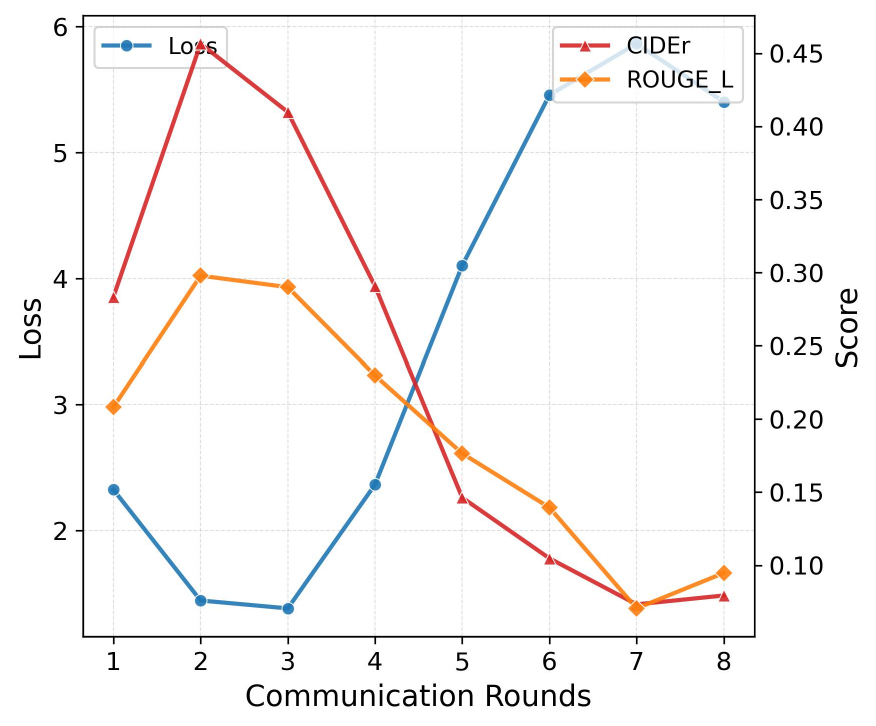}
        \caption{} 
    \end{subfigure}
    \hfill
    \begin{subfigure}{0.48\linewidth}
        \centering
        \includegraphics[width=\linewidth]{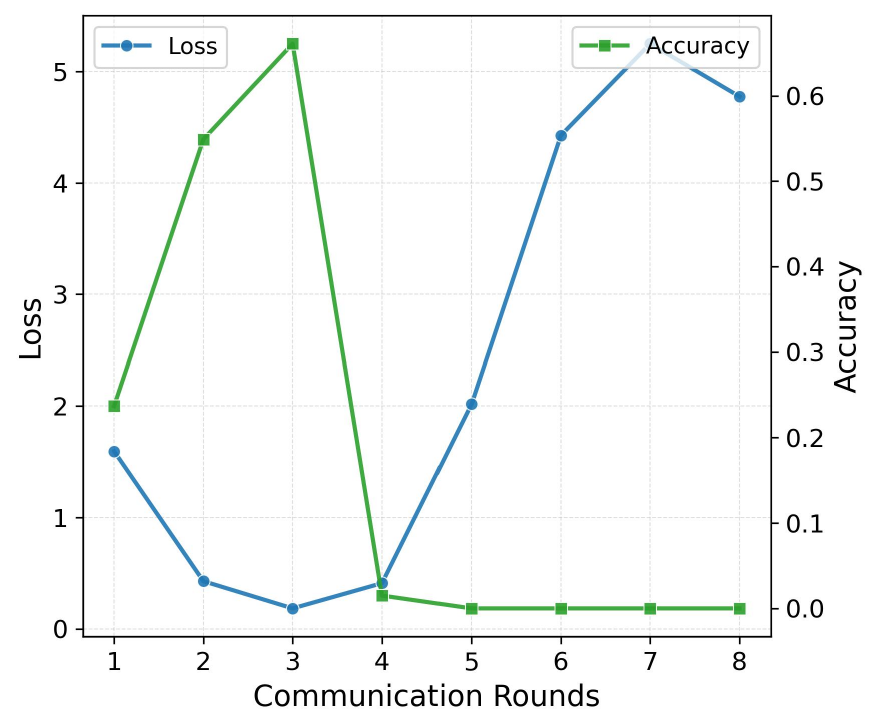}
        \caption{} 
    \end{subfigure}
    \caption{Loss curves and performance trends of FedAdam on caption generation (a) and VQA (b) under Scenario 2 (Fed-Nature).} 
    \label{fig:4}
\end{figure}

\subsection{Convergence and Stability Analysis}
Figure~\ref{fig:convergence} shows the convergence and performance trends of our method and several baselines under Scenario 1. From the loss curves, all methods converge and effectively fit the local data. Although VLM training in our framework starts from the 5th epoch, it reaches a convergence level comparable to the baselines, indicating faster convergence. The performance curves further show that our method achieves better average performance across tasks with improved training stability.
We further observe that FedAdam exhibits severe training instability under Scenario 2. As shown in Figure~\ref{fig:4}, the loss and performance on the VQA and caption generation tasks initially improve, but after several communication rounds, the loss increases sharply while performance drops substantially. This suggests that conventional federated learning algorithms may struggle under complex heterogeneous settings. In contrast, our method maintains stable convergence across diverse heterogeneous scenarios.

\subsection{Different Rank and Efficiency}
We further investigate the impact of different LoRA ranks on model performance. As shown in Table~\ref{tab:lora_rank}, increasing the rank generally benefits generation-oriented tasks. Specifically, caption generation achieves the best performance with rank = 32, while visual grounding performs best with rank = 16. In contrast, VQA achieves its highest accuracy with rank = 8, and classification varies only marginally across different ranks. These results suggest that larger LoRA ranks do not consistently improve performance across all tasks. We further evaluate heterogeneous rank settings by assigning different LoRA ranks to the two VLMs. Although heterogeneous rank settings slightly degrade performance on some tasks compared with homogeneous settings, the proposed framework maintains competitive average performance, demonstrating its robustness to rank heterogeneity.
The communication cost increases with the LoRA rank, reaching approximately 50, 90, and 180 MB (0.72\%, 1.08\%, and 1.79\%) for LLaVA, and 30, 50, and 110 MB (0.46\%, 0.91\%, and 1.80\%) for Show-O at ranks 8, 16, and 32, respectively.
Although the MDA introduces additional communication overhead, it remains lightweight(0.4\%), making the overall communication cost comparable to existing LoRA-based baselines. As the LoRA rank increases, communication cost rises accordingly, while the corresponding performance gains remain limited. These results highlight the trade-off between communication efficiency and performance, indicating that a relatively small LoRA rank is sufficient to achieve competitive average performance.

\subsection{Different Hidden Dimension Size}
We further study the impact of the hidden dimension size $d_h$
in the MDA. As shown in Table \ref{tab:hidden_dim}, $d_h$= 2048 achieves the best overall performance, obtaining the highest accuracy on VQA (0.743), visual grounding (0.519 IoU), and classification (0.918), while maintaining competitive caption generation performance. Reducing $d_h$ to 1024 limits the representation capacity of the MDA, resulting in lower performance on most tasks. In contrast, increasing $d_h$ to 2560 does not provide consistent gains and instead degrades performance on VQA, visual grounding, and classification, suggesting that a larger hidden dimension may not improve knowledge transfer due to increased optimization difficulty. Overall, $d_h$ = 2048 provides a better balance between representation capacity and optimization, and is adopted in all experiments.

\begin{table}[t]
\centering
\small
\setlength{\tabcolsep}{3pt}
\begin{tabular*}{\columnwidth}{@{\extracolsep{\fill}}ccccc}
\toprule
\textbf{$d_h$} &
\textbf{VQA} &
\textbf{Caption Generation} &
\textbf{Visual Grounding} &
\textbf{CLS} \\
\cmidrule(lr){2-5}
& Acc$\uparrow$ & CIDEr/ROUGE-L$\uparrow$ & IoU$\uparrow$ & Acc$\uparrow$ \\
\midrule
1024 & 0.726 & 0.783/0.363 & 0.499 & 0.910 \\
2048 & \textbf{0.743} & \textbf{0.784/0.358} & \textbf{0.519} & \textbf{0.918} \\
2560 & 0.718 & 0.778/0.366 & 0.478 & 0.910 \\
\bottomrule
\end{tabular*}
\caption{Impact of hidden dimension on downstream tasks.}
\label{tab:hidden_dim}
\end{table}

\section{Conclusion}
In this paper, we presented UniFed-VLM, a unified federated instruction tuning framework for vision-language models under multi-dimensional heterogeneity across tasks, modalities, and model architectures. Experimental results on two heterogeneous federated benchmarks show that UniFed-VLM achieves better average performance across tasks compared with representative baselines. These results indicate that combining compensated subspace aggregation with collaborative distillation provides an effective approach for facilitating parameter aggregation and knowledge transfer under heterogeneous tasks, modalities, and model architectures in federated VLMs.

%\section{Acknowledgments}
\bibliography{paper}

\myappendix
\section{Implementation Details}
In our experiments, we perform a total of 20 communication rounds, where each local model is trained for one epoch per round. During the first 5 rounds, we only conduct the first-stage distillation, i.e., the VLM is frozen and distilled into the DMA.

For the MDA architecture, we set the number of layers for $L_{\text{in}}^k$ and $L_{\text{out}}^k$ to 3, respectively, and $L_{\text{att}}^k$ is also configured with 3 layers. Since we adopt LLaMA 3.2-3B and Show-O 1.5B, whose hidden dimensions are 3072 and 2048, respectively, we set the hidden dimension of $\mathrm{LattK}$ to $d_h = 2048$.

For distillation, the temperature for both stages is set to 2.0. The distillation weights are set to $\lambda_{S1} = 0.7$ and $\lambda_{S2} = 0.2$. The second-stage weight is intentionally set smaller to avoid interfering with the task loss.
The number of experts in the MoE module is determined by the number of heterogeneous clients. The learning rate of the gating network is initialized to $1 \times 10^{-4}$ and decays over communication rounds, and $\alpha$=0.4.

We introduce two loss terms for MoE: $\mathcal{L}_{\text{task}}^{\text{MoE}}$ and $\mathcal{L}_{\text{bal}}^{\text{MoE}}$. The task loss $\mathcal{L}_{\text{task}}^{\text{MoE}}$ is identical to the standard autoregressive loss. The balance loss is defined as:
\begin{equation}
\mathcal{L}_{\text{bal}}^{\text{MoE}} =
\frac{1}{E} \sum_{e=1}^{E}
\left( u_e - \frac{1}{E} \right)^2,
\end{equation}
where
\begin{equation}
u_e = \frac{1}{B} \sum_{b=1}^{B} P_{b,e},
\end{equation}
denotes the average utilization of expert $e$, $B$ is the number of samples, and $E$ is the number of experts.
We set the weight of the balance loss to $\lambda_{\text{MoE}} = 0.01$, and the weight of the task loss to $\lambda_{\text{task}}^{\text{MoE}} = 0.3$.

\section{Derivation of the Compensation Objective}
To preserve key information in client-side low-rank updates without significantly increasing communication and storage costs, we adopt separate aggregation of low-rank factors. For layer $l$, the low-rank update of client $i$ is written as
\begin{equation}
\Delta W_i^l = B_i^l A_i^l,
\label{eq:1}
\end{equation}
where $B_i^l \in \mathbb{R}^{d_l \times r_l}$ and $A_i^l \in \mathbb{R}^{r_l \times k_l}$, with $r_l \ll \min(d_l, k_l)$. Compared to directly aggregating the full parameter matrix $\Delta W_i^l$, separately aggregating $A_i^l$ and $B_i^l$ significantly reduces storage and communication costs. However, since the bilinear coupling structure of low-rank factors is decoupled, this aggregation may introduce information loss. To address this issue, we further introduce compensation parameters $\Delta B^l$ and $\Delta A^l$, and refine the global low-rank representation as
\begin{equation}
\widetilde{\Delta W}_m^l =
(\hat B^l + \Delta B^l)(\hat A^l + \Delta A^l),
\label{eq:2}
\end{equation}
where $\hat B^l$ and $\hat A^l$ denote the aggregated global low-rank factors.

\textbf{Motivation from Loss Consistency.} Our goal is to make the compensated global update approximate the client-side updates obtained from local training. Let $\mathcal{L}_i(\cdot)$ denote the local loss of client $i$ at layer $l$. We consider the following loss consistency objective:
\begin{equation}
\mathcal{J} =
\sum_{l=1}^{L}\sum_{i=1}^{N}
\left[
\mathcal{L}_i(W^l + \Delta W_i^l)
-
\mathcal{L}_i(W^l + \widetilde{\Delta W}_m^l)
\right]^2.
\label{eq:3}
\end{equation}
This objective measures the discrepancy between the compensated global update $\widetilde{\Delta W}_m^l$ and the original client update $\Delta W_i^l$ at the loss level. A smaller discrepancy indicates that the global model better approximates the performance of local models.

\textbf{Theorem 1 (First-order Taylor Expansion for Matrix Perturbation).}
Assume $\mathcal{L}_i(W)$ is twice continuously differentiable in a neighborhood of $W^l$. For a sufficiently small perturbation $\Delta W$, we have
\begin{equation}
\mathcal{L}_i(W^l + \Delta W)
=
\mathcal{L}_i(W^l)
+
\langle \nabla \mathcal{L}_i(W^l), \Delta W \rangle
+ O(\|\Delta W\|_F^2),
\label{eq:4}
\end{equation}
where $\langle X, Y \rangle = \mathrm{tr}(X^\top Y)$.
Thus, the first-order approximation is given by
\begin{equation}
\mathcal{L}_i(W^l + \Delta W)
\approx
\mathcal{L}_i(W^l)
+
\langle \nabla \mathcal{L}_i(W^l), \Delta W \rangle.
\label{eq:5}
\end{equation}

\begin{figure*}[h]
    \centering
    \begin{subfigure}{0.32\textwidth}
        \centering
        \includegraphics[width=\linewidth]{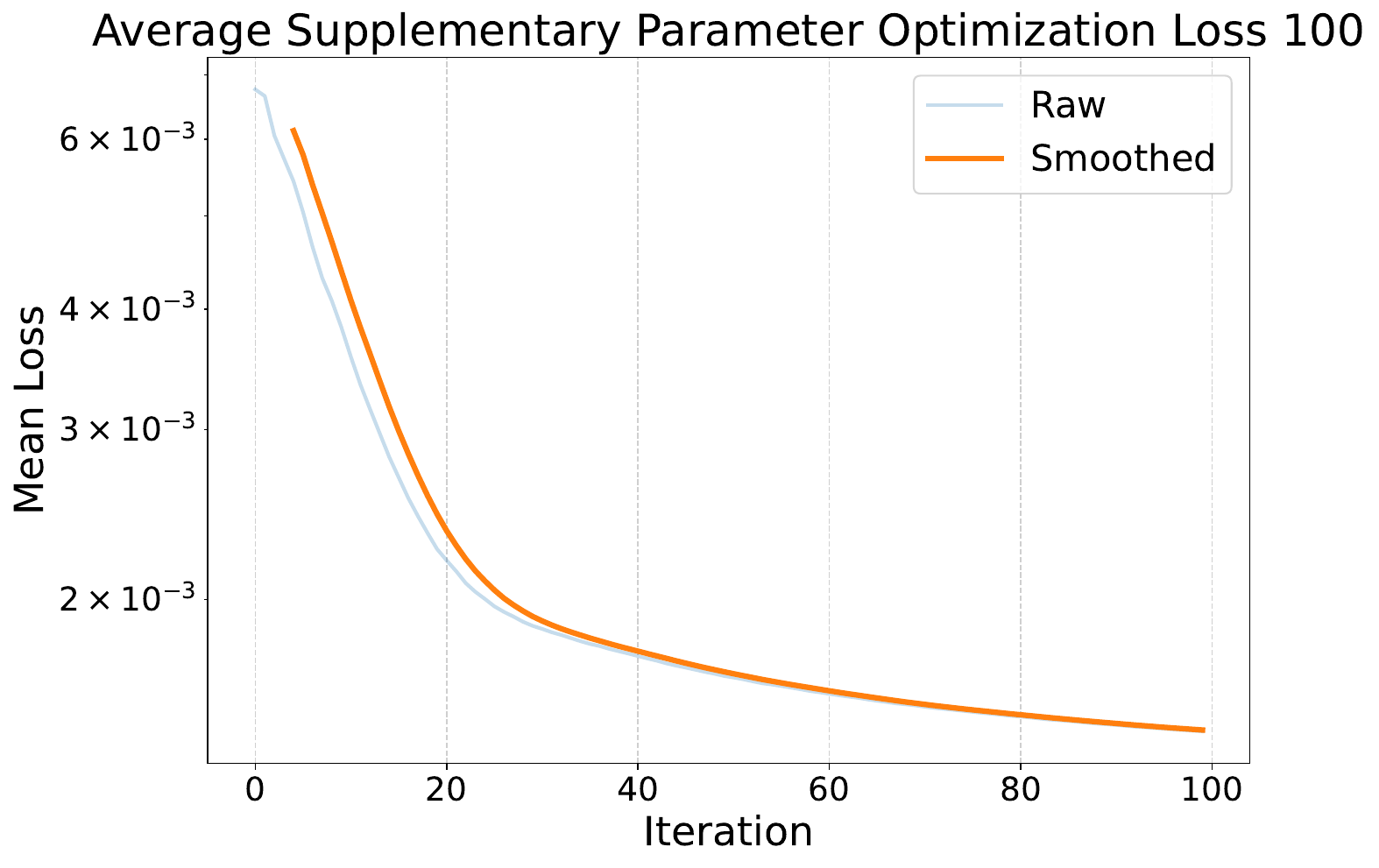}
        \caption{}
    \end{subfigure}
    \hfill
    \begin{subfigure}{0.32\textwidth}
        \centering
        \includegraphics[width=\linewidth]{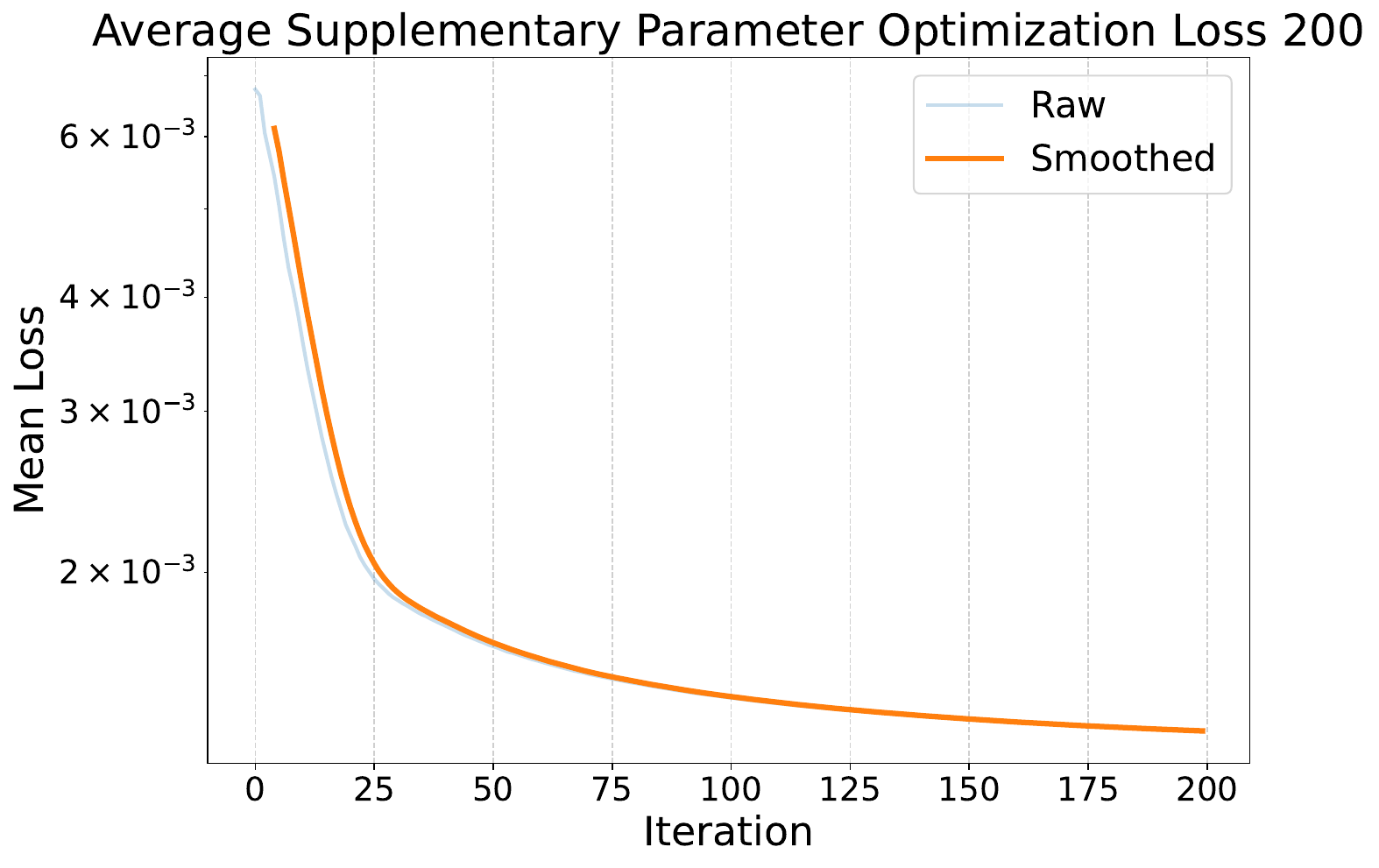}
        \caption{}
    \end{subfigure}
    \hfill
    \begin{subfigure}{0.32\textwidth}
        \centering
        \includegraphics[width=\linewidth]{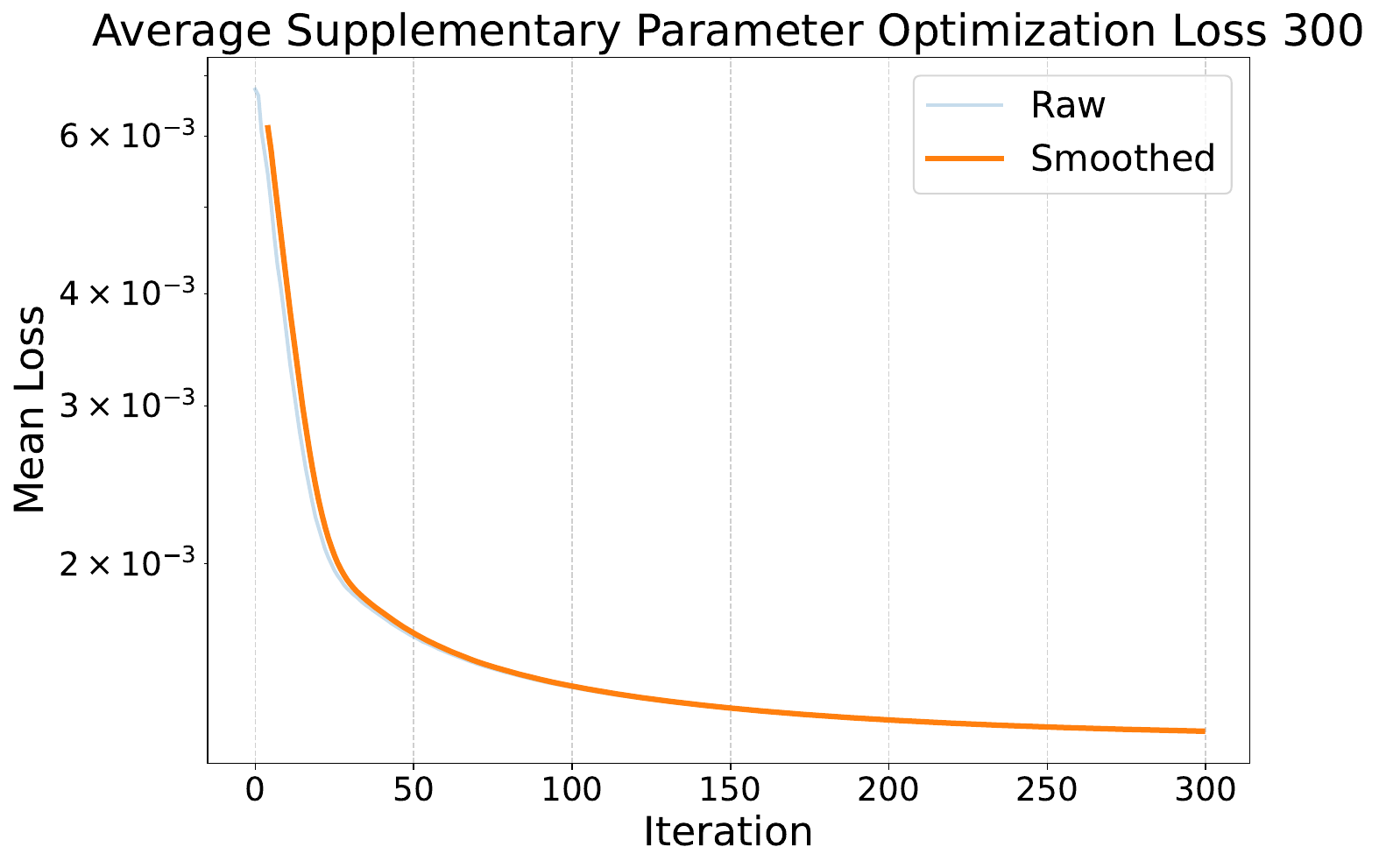}
        \caption{}
    \end{subfigure}
    \caption{Loss curves of the supplementary parameter optimization process under different numbers of iterations for the Show-O model on the Fed-Nature dataset.}
    \label{fig:1}
\end{figure*}

Applying Eq.~\eqref{eq:5} to $\widetilde{\Delta W}_m^l$ and $\Delta W_i^l$, we obtain
\begin{equation}
\mathcal{L}_i(W^l + \widetilde{\Delta W}_m^l)
\approx
\mathcal{L}_i(W^l)
+
\langle \nabla \mathcal{L}_i(W^l), \widetilde{\Delta W}_m^l \rangle,
\label{eq:6}
\end{equation}
\begin{equation}
\mathcal{L}_i(W^l + \Delta W_i^l)
\approx
\mathcal{L}_i(W^l)
+
\langle \nabla \mathcal{L}_i(W^l), \Delta W_i^l \rangle.
\label{eq:7}
\end{equation}
Subtracting the two expressions yields
\begin{equation}
\mathcal{L}_i(W^l + \Delta W_i^l)
-
\mathcal{L}_i(W^l + \widetilde{\Delta W}_m^l)
\approx
\langle
\nabla \mathcal{L}_i(W^l),\,
\Delta W_i^l - \widetilde{\Delta W}_m^l 
\rangle.
\label{eq:8}
\end{equation}
Therefore, Eq.~\eqref{eq:3} can be approximated as
\begin{equation}
\mathcal{J}
\approx
\sum_{l=1}^{L}\sum_{i=1}^{N}
\left[
\langle
\nabla \mathcal{L}_i(W^l),\,
\Delta W_i^l - \widetilde{\Delta W}_m^l
\rangle
\right]^2.
\label{eq:9}
\end{equation}

\textbf{Relation Between Gradient and Local Update.}
During local training, the local update can be viewed as the accumulated effect of gradients along the optimization trajectory. Specifically, the relationship between the local update and gradients can be expressed as:
\begin{equation}
W_{t+1} = W_t - \eta \nabla \mathcal{L}_i(W_t),
\label{eq:10}
\end{equation}
thus the local update can be written as
\begin{equation}
\Delta W_i^l \approx -\eta \nabla \mathcal{L}_i(W^l).
\label{eq:11}
\end{equation}
Equivalently,
\begin{equation}
\nabla \mathcal{L}_i(W^l) \approx -\kappa \Delta W_i^l, \quad \kappa = \frac{1}{\eta}.
\label{eq:12}
\end{equation}
Substituting Eq.~\eqref{eq:12} into Eq.~\eqref{eq:9}, we obtain
\begin{equation}
\mathcal{J}
\approx
\sum_{l=1}^{L}\sum_{i=1}^{N}
\left[
\langle
-\kappa \Delta W_i^l,\,
\Delta W_i^l - \widetilde{\Delta W}_m^l 
\rangle
\right]^2.
\label{eq:13}
\end{equation}
Substituting $\Delta W_i^l = B_i^l A_i^l$ and $\widetilde{\Delta W}_m^l = (\hat B^l + \Delta B^l)(\hat A^l + \Delta A^l)$, we obtain the final objective (Constant $\kappa$  terms can be ignored):
\begin{equation}
\mathcal{L}
=\sum_{l=1}^L\!\sum_{i=1}^N\!
\left[\left\langle
-B_i^lA_i^l,\,
B_i^lA_i^l-\left(\hat{B}^l+\Delta B^l\right)
\left(\hat{A}^l+\Delta A^l\right)
\right\rangle\right]^2.
\label{eq:14}
\end{equation}

\textbf{Interpretation of the Compensation Objective.}The meaning of Eq.~\eqref{eq:14} can be understood from two perspectives.

First, $(\hat B^l, \hat A^l)$ provide the shared low-rank subspace obtained by server-side aggregation, while $(\Delta B^l, \Delta A^l)$ compensate for the lost interaction information caused by factor-wise aggregation. Since low-rank updates inherently rely on the coupling between left and right factors, independently averaging $A_i^l$ and $B_i^l$ breaks this coupling, and the compensation terms restore it.

Second, from an optimization perspective, Eq.~\eqref{eq:14} can be viewed as performing error recovery in the low-rank space. Instead of reconstructing the full high-dimensional update matrix, it searches for a better global representation in the low-dimensional factor space, making it more efficient and suitable for large-scale federated settings.

\textbf{Optimization Procedure.}
In practice, we fix $\{B_i^l, A_i^l\}$ and $\{\hat B^l, \hat A^l\}$, and only optimize the compensation parameters $\Delta B^l$ and $\Delta A^l$. We use Adam as the optimizer and perform a few hundred iterations to obtain stable compensation. The final global factors are
\begin{equation}
B_{\text{global}}^l = \hat B^l + \Delta B^l, \quad
A_{\text{global}}^l = \hat A^l + \Delta A^l,
\end{equation}
and the corresponding global update matrix is
\begin{equation}
\Delta W_{\text{global}}^l =
B_{\text{global}}^l A_{\text{global}}^l.
\end{equation}

This step can be interpreted as a post-processing correction for SVD/low-rank aggregation error, effectively mitigating the information loss introduced by factor-wise aggregation and improving the approximation of client behaviors by the global model.

\section{Analysis of the Number of Iterations}

\begin{table}[t]
\centering
\small
\setlength{\tabcolsep}{3pt}
\begin{tabular*}{\columnwidth}{@{\extracolsep{\fill}}ccccc}
\toprule
\textbf{Iter.} &
\textbf{VQA} &
\textbf{Caption Generation} &
\textbf{Visual Grounding} &
\textbf{CLS} \\
\cmidrule(lr){2-5}
& Acc$\uparrow$ & CIDEr/ROUGE-L$\uparrow$ & IoU$\uparrow$ & Acc$\uparrow$ \\
\midrule
100 & 0.724 & 0.782/0.363 & 0.504 & 0.912 \\
200 & \textbf{0.743} & 0.784/0.358 & \textbf{0.519} & \textbf{0.918} \\
300 & 0.737 & \textbf{0.792/0.368} & 0.497 & 0.910 \\
\bottomrule
\end{tabular*}
\caption{Effect of different numbers of iterations for supplementary parameters on performance on Fed-Nature.}
\label{tab:itr}
\end{table}

As shown in Table~\ref{tab:itr}, we conduct experiments on the Fed-Nature dataset to study the effect of different numbers of iterations for supplementary parameter optimization. Specifically, we set the number of iterations as $I \in \{100, 200, 300\}$. 

From the results, we observe that when $I = 100$, the model performance on all tasks is consistently lower than that with $I = 200$ (the setting adopted in the main paper). When increasing the number of iterations to $I = 300$, the model shows improvement on the Caption Generation task, while its performance on other tasks is inferior to that with $I = 200$.

Furthermore, as illustrated in Fig.~\ref{fig:1}, the loss curve at $I = 100$ (Fig.~\ref{fig:1} (a)) has not fully converged. In contrast, by comparing the loss curves at $I = 200$ (Fig.~\ref{fig:1} (b)) and $I = 300$ (Fig.~\ref{fig:1} (c)), we observe that the model has already largely converged at $I = 200$, with only marginal differences compared to $I = 300$.

Overall, considering both performance and computational efficiency, $I = 200$ provides the best trade-off and is selected as the optimal setting.

\begin{table}[t]
\centering
\small
\setlength{\tabcolsep}{3pt}
\begin{tabular*}{\columnwidth}{@{\extracolsep{\fill}}ccccc}
\toprule
\textbf{$\lambda_{\text{task}}^{MoE}$} &
\textbf{VQA} &
\textbf{Caption Generation} &
\textbf{Visual Grounding} &
\textbf{CLS} \\
\cmidrule(lr){2-5}
& Acc$\uparrow$ & CIDEr/ROUGE-L$\uparrow$ & IoU$\uparrow$ & Acc$\uparrow$ \\
\midrule
0.1 & 0.730 & 0.778/0.355 & 0.520 & 0.919 \\
0.2 & 0.710 & 0.787/0.363 & 0.535 & 0.911 \\
\textbf{0.3} & 0.743 & 0.784/0.358 & 0.519 & 0.918 \\
0.4 & 0.706 & 0.783/0.364 & 0.524 & 0.906 \\
\bottomrule
\end{tabular*}
\caption{Effect of the MoE task loss weight $\lambda_{\text{task}}^{MoE}$.}
\label{tab:moe_weight}
\end{table}

\begin{table}[t]
\centering
\small
\setlength{\tabcolsep}{3pt}
\begin{tabular*}{\columnwidth}{@{\extracolsep{\fill}}ccccc}
\toprule
\textbf{$\alpha$} &
\textbf{VQA} &
\textbf{Caption Generation} &
\textbf{Visual Grounding} &
\textbf{CLS} \\
\cmidrule(lr){2-5}
& Acc$\uparrow$ & CIDEr/ROUGE-L$\uparrow$ & IoU$\uparrow$ & Acc$\uparrow$ \\
\midrule
0.3 & 0.738 & 0.762/0.358 & 0.526 & 0.912 \\
\textbf{0.4} & 0.743 & 0.784/0.358 & 0.519 & 0.918 \\
0.5 & 0.726 & 0.782/0.360 & 0.529 & 0.916 \\
0.6 & 0.724 & 0.781/0.359 & 0.519 & 0.911 \\
\bottomrule
\end{tabular*}
\caption{Effect of coefficient $\alpha$.}
\label{tab:alpha}
\end{table}

\begin{table}[t]
\centering
\small
\setlength{\tabcolsep}{3pt}
\begin{tabular*}{\columnwidth}{@{\extracolsep{\fill}}ccccc}
\toprule
\textbf{Layers} &
\textbf{VQA} &
\textbf{Caption Generation} &
\textbf{Visual Grounding} &
\textbf{CLS} \\
\cmidrule(lr){2-5}
& Acc$\uparrow$ & CIDEr/ROUGE-L$\uparrow$ & IoU$\uparrow$ & Acc$\uparrow$ \\
\midrule
2 & 0.694 & 0.776/0.363 & 0.513 & 0.912 \\
\textbf{3} & 0.743 & 0.784/0.358 & 0.519 & 0.918 \\
4 & 0.721 & 0.847/0.374 & 0.516 & 0.907 \\
\bottomrule
\end{tabular*}
\caption{Effect of the number of projection layers in MDA.}
\label{tab:projection_layer}
\end{table}

\section{Hyperparameter Analysis}

\subsection{Effect of MoE-related Hyperparameters}
We first investigate the impact of MoE-related hyperparameters, including the task loss weight $\lambda_{\text{task}}^{MoE}$, the coefficient $\alpha$, and the number of projection layers in MDA. The experiments are conducted on Fed-Nature under Scenario 1, following the same evaluation protocol as the main experiments. These parameters control the balance between task optimization, expert collaboration, and feature transformation capability.

\subsubsection{Impact of $\lambda_{\text{task}}^{MoE}$}
We first investigate the effect of the MoE task loss weight $\lambda_{\text{task}}^{MoE}$, which controls the balance between task-specific optimization and collaborative knowledge learning in the MoE module. As shown in Table \ref{tab:moe_weight}, different values of $\lambda_{\text{task}}^{MoE}$ lead to different trade-offs across tasks. When $\lambda_{\text{task}}^{MoE}$ is set to 0.1 or 0.2, the performance on VQA and CLS decreases, indicating insufficient task supervision from the MoE module. In contrast, a larger weight (e.g., 0.4) causes performance degradation on most tasks, suggesting that excessive emphasis on MoE optimization may interfere with the original task objective. The setting of $\lambda_{\text{task}}^{MoE}=0.3$ achieves the best overall balance, obtaining the highest VQA accuracy and competitive performance on other tasks. Therefore, we adopt 0.3 as the default value.

\subsubsection{Impact of $\alpha$}
We further analyze the impact of the coefficient $\alpha$, which regulates the contribution of collaborative knowledge during the optimization process. As reported in Table \ref{tab:alpha}, the model performance varies slightly with different $\alpha$ values. A smaller value ($\alpha=0.3$) limits the effect of knowledge interaction, resulting in lower CLS accuracy. Increasing $\alpha$ to 0.5 improves visual grounding performance but slightly reduces VQA accuracy, indicating that excessive knowledge transfer may introduce additional interference among heterogeneous clients. The setting of $\alpha=0.4$ achieves a better balance between knowledge collaboration and task-specific adaptation, leading to the best overall performance. Thus, we select $\alpha=0.4$ in our experiments.

\subsubsection{Impact of Projection Layer Depth}
We investigate the influence of the projection layer depth in MDA to evaluate its effect on feature transformation capability. As shown in Table \ref{tab:projection_layer}, increasing the number of projection layers from 2 to 3 consistently improves performance across most tasks, demonstrating that deeper transformations provide more effective feature alignment between heterogeneous models. However, further increasing the depth to 4 does not lead to consistent improvements. Although it benefits caption generation, it slightly degrades VQA and classification performance, likely due to increased optimization difficulty and unnecessary feature transformation. Therefore, we choose 3 projection layers as the default configuration, which provides a favorable balance between representation capacity and optimization stability.

\subsection{Effect of Distillation Hyperparameters}
We further investigate the impact of distillation hyperparameters in the two-stage collaborative distillation process. Specifically, we analyze the effects of $\lambda_{S1}$ and $\lambda_{S2}$, which control the knowledge transfer strength from VLMs to MDA and from MDA back to VLMs, respectively. The experiments are conducted on Fed-Nature under Scenario 1 following the same evaluation protocol as the main experiments.
\begin{table}[t]
\centering
\small
\setlength{\tabcolsep}{3pt}
\begin{tabular*}{\columnwidth}{@{\extracolsep{\fill}}ccccc}
\toprule
\textbf{$\lambda_{S1}$-$\lambda_{S2}$} &
\textbf{VQA} &
\textbf{Caption Generation} &
\textbf{Visual Grounding} &
\textbf{CLS} \\
\cmidrule(lr){2-5}
& Acc$\uparrow$ & CIDEr/ROUGE-L$\uparrow$ & IoU$\uparrow$ & Acc$\uparrow$ \\
\midrule
0.7-0.1 & 0.729 & 0.777/0.355 & 0.513 & 0.912 \\
0.7-0.3 & 0.739 & 0.729/0.352 & 0.515 & 0.918 \\
\textbf{0.7-0.2} & 0.743 & 0.784/0.358 & 0.519 & 0.918 \\
0.6-0.2 & 0.727 & 0.788/0.362 & 0.520 & 0.914 \\
0.8-0.2 & 0.722 & 0.821/0.371 & 0.517 & 0.905 \\
\bottomrule
\end{tabular*}
\caption{Effect of distillation weights $\lambda_{S1}$ and $\lambda_{S2}$.}
\label{tab:lambda_s}
\end{table}

\subsubsection{Impact of $\lambda_{S1}$ and $\lambda_{S2}$}
As shown in Table~\ref{tab:lambda_s}, the two distillation weights jointly influence the trade-off between cross-model knowledge transfer and task-specific optimization. When fixing $\lambda_{S1}=0.7$, increasing $\lambda_{S2}$ from 0.1 to 0.2 consistently improves performance across different tasks, demonstrating that appropriate feedback distillation facilitates knowledge transfer between heterogeneous VLMs. However, further increasing $\lambda_{S2}$ to 0.3 leads to performance degradation, particularly in caption generation, where CIDEr decreases from 0.784 to 0.729. This suggests that excessive feedback distillation may introduce over-regularization, limiting the model's ability to adapt to task-specific characteristics. For $\lambda_{S1}$, a smaller weight limits the initial knowledge transfer from VLMs to MDA, while a larger weight overemphasizes the first-stage alignment and causes performance degradation on discriminative tasks such as classification. The combination of $\lambda_{S1}=0.7$ and $\lambda_{S2}=0.2$ achieves a better balance between the two distillation stages, obtaining consistent performance across different tasks and is therefore adopted as the default setting.

\section{Another Scenario}
To further evaluate the robustness of UniFed-VLM under different task-model assignments, we further introduce another heterogeneous scenario, where LLaVA is responsible for visual question answering (VQA) and visual grounding, while Show-O handles classification and caption generation. This setting involves different task distributions across heterogeneous VLMs, requiring effective collaboration among models with distinct capabilities.
As shown in Table~\ref{tab:Scenario}, UniFed-VLM achieves competitive average performance across all tasks. Specifically, it obtains the best results on VQA, visual grounding, and classification, achieving 0.738 Acc, 0.365 IoU, and 0.899 Acc, respectively. For caption generation, FedAvg achieves slightly better performance, while UniFed-VLM remains comparable with other methods. These results indicate that UniFed-VLM can effectively balance task-specific adaptation and cross-model knowledge transfer under different heterogeneous configurations.
Compared with other federated fine-tuning methods, FedAvg and FedProx are limited by direct parameter aggregation, while HetLoRA and pFedLoRA still face challenges in transferring complementary knowledge across heterogeneous VLMs. In contrast, UniFed-VLM combines FedCSA and TCoD to improve intra-architecture aggregation and cross-model collaboration, resulting in more consistent performance across diverse tasks.
Overall, this additional scenario demonstrates the generality of UniFed-VLM under different task-model assignments and validates its effectiveness in addressing multi-dimensional heterogeneity in federated VLM adaptation.

\begin{table}[t]
\centering
\small
\setlength{\tabcolsep}{3pt}
\begin{tabular*}{\columnwidth}{@{\extracolsep{\fill}}ccccc}
\toprule
\textbf{Method} &
\textbf{VQA} &
\textbf{Caption Generation} &
\textbf{VG} &
\textbf{CLS} \\
\cmidrule(lr){2-5}
& Acc$\uparrow$ & CIDEr/ROUGE-L$\uparrow$ & IoU$\uparrow$ & Acc$\uparrow$ \\
\midrule
FedAvg & 0.725 & \textbf{0.777/0.348} & 0.347 & 0.892 \\
FedProx & 0.716 & 0.738/0.344 & 0.355 & 0.888 \\
HetLoRA & 0.717 & 0.766/0.342 & 0.342 & 0.891 \\
pFedLoRA & 0.571 & 0.665/0.324 & 0.307 & 0.837 \\
\textbf{Ours} & \textbf{0.738} & 0.765/0.346 & \textbf{0.365} & \textbf{0.899} \\
\bottomrule
\end{tabular*}
\caption{Evaluation under another heterogeneous task-model scenario on Fed-Nature.}
\label{tab:Scenario}
\end{table}

\begin{table}[t]
\centering
\small
\setlength{\tabcolsep}{3pt}
\begin{tabular*}{\columnwidth}{@{\extracolsep{\fill}}ccccc}
\toprule
\textbf{Method} &
\textbf{VQA} &
\textbf{Caption Generation} &
\textbf{VG} &
\textbf{CLS} \\
\cmidrule(lr){2-5}
& Acc$\uparrow$ & CIDEr/ROUGE-L$\uparrow$ & IoU$\uparrow$ & Acc$\uparrow$ \\
\midrule
\textbf{Ours} & 0.743 & 0.784/0.358 & 0.519 & 0.918 \\
Ours (w/ DP) & 0.724 & 0.781/0.360 & 0.521 & 0.918 \\
\bottomrule
\end{tabular*}
\caption{Privacy-preserving evaluation of UniFed-VLM on Fed-Nature under Scenario 1. Differential privacy (DP) is applied to LoRA updates before federated aggregation.}
\label{tab:privacy}
\end{table}

\begin{algorithm}[t]
\small
\caption{UniFed-VLM}
\label{alg:unifed}
\begin{algorithmic}[1]

\REQUIRE 
$K$ clients with datasets $\{D_k\}_{k=1}^K$, heterogeneous VLMs $\{M_k\}$, 
federated rounds $T$, local epochs $E$, hyperparameters $\lambda_{\text{S1}},\lambda_{\text{S2}},\lambda_{\text{MoE}},\beta,\alpha$...

\ENSURE Fine-tuned heterogeneous VLMs $\{M_k\}_{k=1}^K$

\STATE \textbf{Initialization}: Global LoRA params $\{\hat{A}^l,\hat{B}^l\}_{l=1}^L$, shared MDA attention $L_{\text{att}}$, momentum buffers $v_{A,i}^l,v_{B,i}^l$ for all layers $l=1,\dots,L$ 
\FOR{$t = 1, 2, \dots, T$}

    \FOR{each client $k$ \textbf{in parallel}}

        \STATE Receive global LoRA $\{\hat{A}^l,\hat{B}^l\}_{l=1}^L$, MDA attention $L_{\text{att}}$ and MoE expert pool from server. Then load parameters  into model

        \STATE \textcolor{gray}{ToCD: Two-stage Collaborative Distillation}

        \FOR{$e = 1, \dots, E$}

            \STATE \textcolor{gray}{Stage 1: VLM $\rightarrow$ MDA Distillation}

            \STATE Freeze $M_k$, train MDA $f_k = L_{\text{out}}^k(L_{\text{att}}(L_{\text{in}}^k(x)))$ via:

            \STATE $\quad \mathcal{L}_{\text{Stage1}} = \mathcal{L}_{\text{task}}^l + \lambda_{\text{S1}} \mathcal{L}_{\text{KD}}(p_l \parallel p_V)$

            \STATE $\quad$ where $\mathcal{L}_{\text{KD}} = \frac{1}{|M|}\sum_{i\in M} D_{\text{KL}}(p_i^{(s)} \parallel p_i^{(t)})$

            \STATE \textcolor{gray}{Stage 2: MDA (MoE) $\rightarrow$ VLM Distillation}

           \STATE Freeze MDA modules and train the MoE gating network to compute weights $g$; fuse Top-$K$ experts with the anchor, then optimize LoRA parameters via:

            \STATE $\displaystyle \mathcal{L}_{\text{Stage2}} =
                \mathcal{L}_{\text{task}}^V
                + \lambda_{\text{S2}}\mathcal{L}_{\text{KD}}(p_V \parallel p_l)
                + \lambda_{\text{task}}^{\text{MoE}}\mathcal{L}_{\text{task}}^{\text{MoE}}
                + \lambda_{\text{MoE}}\mathcal{L}_{\text{Bal}}^{\text{MoE}}$

        \ENDFOR

        \STATE Upload local LoRA $\{A_k^l, B_k^l\}_{l=1}^L$ and MDA attention $L_{\text{att}}^k$ to server

    \ENDFOR

    \STATE \textcolor{gray}{FedCSA: Federated Compensated Subspace Aggregation}

    \FOR{$G$(model architecture group), each layer $l$}

        \STATE Concatenate LoRA$\{A_i^l, B_i^l\}_{i=1}^G$ \& apply randomized SVD:
        \STATE $\quad A_f^l = U_A^l \Sigma_A^l (V_A^l)^\top$, $B_f^l = U_B^l \Sigma_B^l (V_B^l)^\top$

        \STATE Partition $V_A^l, U_B^l$ to get task features $\{V_i^l\}_{i=1}^G,\{U_i^l\}_{i=1}^G$

        \STATE Compute dynamic weights $w_{A,i}^l$ via momentum on update magnitude (same for $w_{B,i}^l$)

        \STATE Aggregate subspace features:
        \STATE $\quad \widetilde{V}^l = \sum_{i\in G} w_{A,i}^l V_i^l$, $\widetilde{U}^l = \sum_{i\in G} w_{B,i}^l U_i^l$

        \STATE Reconstruct \& compensate global LoRA:

        \STATE $\quad \hat{A}^l = U_A^l \Sigma_A^l (\widetilde{V}^l)^\top + \Delta A^l$

        \STATE $\quad \hat{B}^l = \widetilde{U}^l \Sigma_B^l V_B^l + \Delta B^l$

    \ENDFOR

    \STATE \textcolor{gray}{TCoD: Construct MoE MDA}
    
    \STATE \textbf{Update MoE experts}: 
    \FOR{each client $k=1,\dots,K$}
    \STATE Update the MoE expert pool of client $k$: 
   $\{E_i\}_{i=1}^N\leftarrow \{L_{\text{att}}^k\}_{k=1}^N$,
    $N$ denotes the number of clients with architectures heterogeneous to the current client, including itself.
    \ENDFOR
    
    \STATE Broadcast updated global LoRA $\{\hat{A}^l, \hat{B}^l\}_{l=1}^L$ and new MoE expert pool $\{E_i\}_{i=1}^N$ to the corresponding client

\ENDFOR

\end{algorithmic}
\end{algorithm}

\section{Privacy Issues Analysis}
To further investigate the impact of privacy preservation on UniFed-VLM, we incorporate differential privacy (DP) into the federated fine-tuning process. Specifically, before performing federated aggregation, Gaussian noise is added to the clipped LoRA updates uploaded by each client to protect sensitive information while maintaining the personalized adaptation capability of local models. As shown in Table~\ref{tab:privacy}, applying DP introduces only limited performance variations compared with the original UniFed-VLM.

Although the injected noise inevitably affects the quality of uploaded updates, UniFed-VLM with DP maintains competitive performance across different tasks. Specifically, VQA accuracy decreases from 0.743 to 0.724, indicating that reasoning-oriented tasks are relatively more sensitive to perturbations. In contrast, caption generation, visual grounding, and classification exhibit only marginal changes, with the performance remaining comparable to the non-private setting. These results demonstrate that UniFed-VLM can effectively support differential privacy while maintaining reliable federated multimodal learning performance, highlighting its compatibility with privacy-preserving mechanisms in practical federated scenarios.

\section{Federated Optimization}
We summarize the overall optimization procedure of UniFed-VLM in Algorithm \ref{alg:unifed}. The framework performs federated optimization over $T$ communication rounds, where each round consists of local two-stage collaborative distillation and server-side aggregation. At the beginning of each round, clients receive the global LoRA parameters and the shared MDA module from the server to initialize their local VLMs.

During local training, each client performs $E$ local epochs with two-stage optimization. In Stage 1, the VLM is frozen, and the MDA module is optimized to learn task-specific representations from the VLM. The MDA is trained using $\mathcal{L}{S1}$, which combines the task objective and knowledge distillation loss from the VLM. In Stage 2, the MDA attention module is fixed, and the LoRA parameters together with the MoE gating network are optimized. The LoRA-adapted VLM learns from the MDA through reverse distillation with $\mathcal{L}{S2}$, which incorporates two task losses, distillation loss, and MoE balancing loss. After local training, clients upload their updated LoRA parameters and MDA attention modules to the server.

On the server side, FedCSA first aggregates LoRA parameters among clients with the same VLM architecture. Specifically, LoRA matrices are transformed into a shared low-rank subspace through randomized SVD, followed by adaptive subspace aggregation based on update stability and lightweight compensation optimization to reduce information loss. Then, for each client, the uploaded MDA attention modules from clients with heterogeneous VLM architectures relative to the local model are utilized to update its MoE expert pool, facilitating cross-model knowledge collaboration while preserving client-specific expertise. Finally, the updated LoRA parameters and MoE experts are returned to the corresponding client for the next round of communication. Through this iterative process, UniFed-VLM enables federated VLM instruction tuning under simultaneous task, modality, and model heterogeneity.

\section{Limitations}
While UniFed-VLM achieves effective federated adaptation of heterogeneous VLMs, several limitations remain. The two-stage collaborative distillation in TCoD introduces additional client-side computation due to extra training stages, which may challenge resource-constrained devices. Moreover, FedCSA incurs extra server-side overhead from SVD decomposition and compensation parameter optimization. In our experiments, FedAvg and FedProx introduce negligible aggregation costs, while FlexLoRA and HetLoRA require approximately 1 minute per communication round. FedCSA takes around 2 minutes due to the additional optimization process, which remains acceptable in our setting but may affect scalability with more participating clients.
Furthermore, TCoD relies on effective knowledge transfer across heterogeneous VLMs, and its performance may degrade when the architectural gap becomes extremely large. Finally, our evaluation focuses on vision-language tasks, and the generalization of UniFed-VLM to broader multimodal scenarios remains to be explored.

\end{document}